%% file: main.tex
\pdfoutput=1
\documentclass{article}

\usepackage[eandd,final]{neurips_2026}

\usepackage[utf8]{inputenc}
\usepackage[T1]{fontenc}
\usepackage{courier}
\usepackage{hyperref}
\hypersetup{hidelinks,pdftitle={Verifier Choice is a Benchmark Design Variable: Auditing Structural Counting Evaluation in Text-to-Image Models}}
\usepackage{url}
\usepackage{booktabs}
\usepackage{amsfonts}
\usepackage{amsmath}
\usepackage{amssymb}
\usepackage{microtype}
\usepackage{xcolor}
\usepackage{graphicx}
\usepackage{subcaption}
\usepackage{multirow}
\usepackage{array}
\usepackage{enumitem}
\usepackage{makecell}
\usepackage{mathtools}
\usepackage{siunitx}

\newcommand{\pp}{\,\mathrm{pp}}
\newcommand{\valid}{\mathrm{valid}}

\newcommand{\longwang}[1]{{{{\textcolor{black}{#1}}}}}

\title{Verifier Choice is a Benchmark Design Variable:\\Auditing Structural Counting Evaluation in Text-to-Image Models}

\input{authors}

\begin{document}

\maketitle

\begin{abstract}
Automatic evaluation is now central to text-to-image benchmarking, but the verifier that converts generated images into scores is often treated as an implementation detail. We argue that this is unsafe: an automatic benchmark score is a property of the image generator, prompt distribution, verifier, verifier-use protocol, and scoring rule, not of the generator alone. We study this issue in structural counting, a controlled setting where human-visible counts can be audited and where both instance-level object counts and parent-bound substructure counts arise naturally. We construct a verifier-centric audit suite over seven counting \longwang{object}s: cell phones, chairs, and traffic lights for object counting, and clover leaflets, dresser drawer fronts, power-strip sockets, and turbine blades for substructure counting. Across 6,240 generated images from Qwen-Image-2512 and FLUX.2-dev, we compare open MLLM verifiers under count-only, reject-aware, zero-shot and few-shot protocols, plus TIFA-like target-query variants and targeted frontier closed-model sanity checks on two hard \longwang{object}s. We also compare MLLMs with a detector baseline. On a balanced 2,280-image evaluation split, aggregate score matching can coexist with image-level disagreement, and verifier-use protocols and generator distributions change the measured reliability. Every verifier--protocol configuration also shrinks the human score difference between the two generators, by $7.5$--$25.0\pp$, and with a third generator, Z-Image, all four tested protocols reverse the human ordering of Z-Image and FLUX.2-dev on our fixed prompt suite. Targeted GPT-5.5 and Claude Opus 4.7 checks do not remove the failure modes: GPT-5.5 still has only 59.92\% clover valid-count fidelity, while Claude Opus 4.7 has lower same-slice agreement ($\kappa=0.432$). We propose reporting automatic benchmark scores with a factorized verifier reliability profile: aggregate score gap, image-level agreement, generator-comparison distortion, valid-image count fidelity, rejection behavior, output compliance, and directional bias.
\end{abstract}

\section{Introduction}
\label{sec:introduction}

Automatic evaluation has become a standard component of text-to-image (T2I) benchmarking. A typical benchmark chooses a target capability, constructs prompts, generates images, applies a verifier, and aggregates the verifier outputs into a score. Existing benchmarks instantiate this pipeline in different ways. TIFA decomposes prompts into visual questions and checks generated images with VQA models \citep{tifa}. GenEval uses an object-focused detection pipeline for co-occurrence, count, position, and color \citep{geneval}. T2I-CompBench++, DSG, VQAScore/GenAI-Bench, HEIM, and Gecko expand automatic or semi-automatic evaluation across compositional, question-answering, holistic, and human-rating settings \citep{t2icompbench,dsg,vqascore,heim,gecko}. These efforts have made T2I evaluation more scalable and more fine-grained. They also make verifier design an important part of the measurement pipeline.

Prior benchmarks often document their verifier choices but usually fix them within a benchmark rather than study them as variables that can change the resulting model conclusions. We argue that this choice should be treated as part of the benchmark itself. A benchmark score is a property of the generator, prompt distribution, verifier, verifier-use protocol, and scoring rule:
\begin{equation}
    S(G,\mathcal{P},V,\pi,\sigma)
    = \mathbb{E}_{p\sim \mathcal{P},\,x\sim G(p)}
      \left[\sigma\bigl(p,x,V_{\pi}(x,p)\bigr)\right].
    \label{eq:score_tuple}
\end{equation}
Here $G$ is the T2I generator, $\mathcal{P}$ is the prompt set or prompt distribution, $V$ is the verifier model family, $\pi$ is the \emph{verifier-use protocol}, and $\sigma$ is the \emph{scoring rule}. The verifier-use protocol includes the query form, prompt wording, few-shot support examples, \longwang{object} card, output schema, parser, detector thresholds, and rejection criteria. The scoring rule specifies how the per-image verifier output is converted into benchmark pass/fail labels and then aggregated. Under this view, different studies can evaluate the same generated images and still reach different conclusions if they use different verifiers, verifier-use protocols, or scoring rules.

\begin{figure}[t]
\centering
\includegraphics[width=\linewidth]{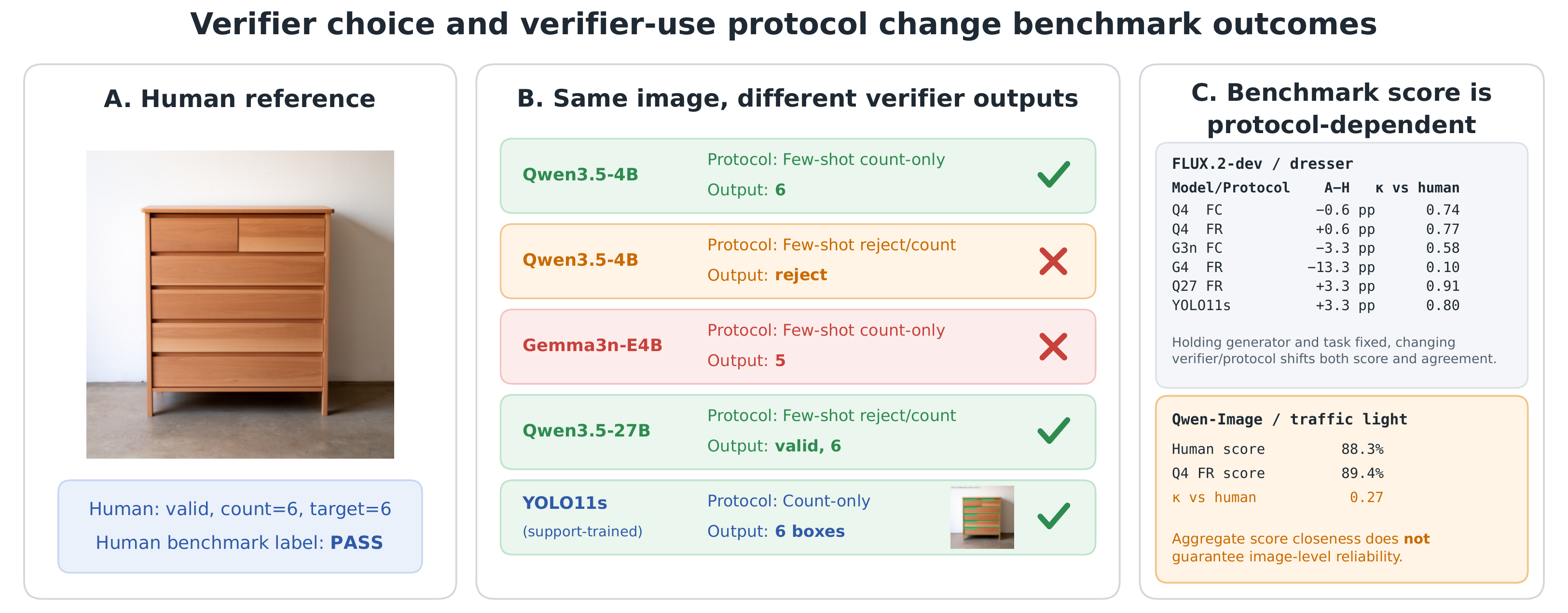}
\caption{
\textbf{Verifier choice and verifier-use protocol change benchmark outcomes.}
Panel A shows a human-audited generated image. Panel B shows that the same image can receive different
outputs under different verifiers and verifier-use protocols. Panel C reports aggregate metrics on the
corresponding generator--\longwang{object} slices (Q4/Q27: Qwen3.5-4B/27B; G3n/G4: Gemma3n-E4B/Gemma4-E4B; FC/FR: few-shot count-only/reject-count), where the generated images and human labels
are fixed but automatic scores and image-level agreement vary across
verifier/protocol choices. This illustrates why a T2I benchmark score should be
reported as a property of the generator, prompt distribution, verifier,
verifier-use protocol, and scoring rule.
}
\label{fig:intro}
\end{figure}

Figure~\ref{fig:intro} illustrates the issue.
\longwang{An image was generated with a prompt for a dresser with six drawers.
For the generated dresser image, humans mark it as valid with a count of six drawers,
but different verifiers and verifier-use protocols produce different outputs:}
one Multimodal Large Language Model (MLLM) returns the correct count under a count-only protocol but rejects the image under a validity-aware protocol, another under-counts, and a support-trained detector counts correctly. These image-level disagreements also appear at the benchmark level. On the corresponding generator--\longwang{object} slice, automatic scores and Cohen's $\kappa$ vary substantially across verifier-use protocols, even though the generated images and human labels are fixed. This motivates treating verifier choice and verifier-use protocol as benchmark-design variables rather than implementation details.

Recent work has begun to audit T2I metrics themselves. DSG identifies reliability failures in question-generation/question-answering based evaluation \citep{dsg}; Gecko studies how metrics, prompts, and human rating templates affect conclusions \citep{gecko}; \citet{whatmakesagoodmetric} show that popular automatic metrics can be insensitive to some visual and linguistic properties and may exploit shortcuts; and GenEval~2 frames judge/prompt mismatch as a form of benchmark drift \citep{geneval2}. We study how the selection of verifier and verifier-protocol impacts the reliability of the text-to-image evaluation.
\longwang{We hold generated images fixed and treat the verifier and verifier-use protocol as the primary experimental variables in this study.}

We study this problem through \longwang{an audit task of \emph{structural counting} for three reasons}. First, counting probes a primitive operation behind many compositional evaluations: converting a visual input into a discrete inventory of target instances or target parts. Many object-relation, attribute-binding, and spatial-composition judgments implicitly depend on such an inventory, since the verifier must decide which visual units exist before judging their relations or attributes. Second, many counting errors are human-visible, so automatic judgments can be checked at the image level. Third, the task admits multiple verifier families: specialist detectors can count some fixed object categories, while \longwang{general} MLLMs can be queried across open-vocabulary objects and object parts.

We therefore build a verifier-centric audit suite with two regimes. In \emph{object counting}, the target is the number of visible instances of a category, such as chairs or cell phones. In \emph{parent-bound substructure counting}, the target is the number of visible subparts of a main object, such as clover leaflets, dresser drawer fronts, power-strip sockets, or turbine blades. The suite contains seven \longwang{object}s, three prompt templates per \longwang{object}, and two T2I generators: Qwen-Image-2512 and FLUX.2-dev \citep{qwenimage,qwenimage2512,flux2}. We generate and human-audit 6,240 images in total. The main MLLM audit evaluates a balanced 2,280-image subset per verifier-use protocol, with image-level bootstrap confidence intervals.

Our audit yields six main findings:
\textbf{(1) aggregate score matching does not imply reliable verification,}
since some verifier--protocol pairs have near-zero score gaps while disagreeing
with humans at the image level;
\textbf{(2) verifier errors distort generator comparisons,}
since every configuration shrinks the human difference between Qwen-Image-2512 and
FLUX.2-dev, and with a third generator all four tested protocols reverse the human
ordering of Z-Image and FLUX.2-dev on the fixed suite;
\textbf{(3) verifier-use protocols are measurement variables,}
because few-shot support, reject/count schemas, target-query formulations,
detector thresholds and even the rendering of multimodal prompts change benchmark
outcomes non-monotonically;
\textbf{(4) detector protocols are useful but regime-specific,}
working best when target units are visually regular and support labels or
threshold tuning are available, but failing on reject-boundary cases;
\textbf{(5) failures are \longwang{object}-local and directional,}
with traffic-light validity shifts and clover/substructure counting errors
creating large \longwang{object}-specific score distortions; and
\textbf{(6) stronger closed MLLMs do not remove these failure modes,}
as frontier-model checks shift the operating point but do not eliminate clover
counting or traffic-light rejection errors. Together, these results support the main recommendation of the paper: automatic T2I benchmark scores should be reported with a factorized verifier reliability profile, including aggregate score gap, image-level agreement, generator-comparison distortion, valid-image count fidelity, rejection behavior, output compliance, and signed count bias.

The audit artifact comprises generated images, prompts, target counts, human annotations, verifier outputs, support-example indices, detector predictions, evaluation scripts, and metadata. The data are released at \url{https://huggingface.co/datasets/neurips2026-ed443/audit-release} and the code at \url{https://github.com/AzusaCat1111/verifier-choice-audit}.

\paragraph{Contributions.}
We formalize automatic T2I benchmark scores as properties of generator, prompt distribution, verifier, verifier-use protocol, and scoring rule, making verifier choice an explicit benchmark-design variable. We introduce a human-audited structural-counting audit suite covering object counts and parent-bound substructure counts across seven target units, three prompt templates, two generators, and 6,240 generated images, and extend the generator comparison with a third generator, Z-Image. We use this suite to answer three verifier-reliability questions: whether aggregate score matching implies image-level agreement and faithful generator comparisons, how verifier-use protocols change measurement, and where failures localize and whether they change generator rankings. We organize these materials as an audit artifact and propose a verifier selection card / reliability profile for reporting future automatic T2I benchmark results.

\section{Related Work}
\label{sec:related-work}

\paragraph{Automatic T2I evaluation.}
Early scalable T2I metrics often measured holistic image--text similarity or image distribution quality. Recent benchmarks instead decompose prompts into finer capabilities. TIFA converts prompts into question--answer pairs and evaluates faithfulness with VQA \citep{tifa}. GenEval focuses on object-level capabilities and uses detection-based checks for object presence, counting, position, and color \citep{geneval}. T2I-CompBench++ evaluates compositional prompts with multiple category-specific metrics and studies the potential of MLLMs for evaluation \citep{t2icompbench}. VQAScore reframes image--text alignment as image-to-text likelihood and introduces GenAI-Bench for compositional prompts \citep{vqascore}. DSG uses scene-graph-structured atomic questions to reduce QG/VQA reliability failures \citep{dsg}. HEIM and Gecko broaden evaluation to many dimensions, including human ratings and metric comparisons \citep{heim,gecko}. We ask how much the verifier and verifier-use protocol themselves change benchmark conclusions on the same audited images.

\paragraph{Auditing evaluation metrics.}
Several recent studies show that automatic T2I metrics are themselves objects of measurement. DSG documents inconsistencies in question-generation/question-answering pipelines \citep{dsg}. Gecko measures how prompt sets, human templates, and automatic metrics alter model comparisons \citep{gecko}. \citet{whatmakesagoodmetric} find that CLIPScore, TIFA, VPEval, and DSG do not satisfy all desiderata for text-image consistency metrics and can be vulnerable to textual or visual shortcuts. GenEval~2 argues that benchmark drift arises when the judge no longer matches model capabilities and prompt difficulty \citep{geneval2}. Our contribution is a controlled, image-level audit of verifier-use choices in a capability where human labels are relatively direct.

\paragraph{Counting as a diagnostic ability.}
Counting is a common T2I stress test because it is easy to state in prompts, appears in object-focused benchmarks, and exposes failures in instance separation, object binding, and structural generation. Counting also has a useful split for verifier auditing. Common object categories can be checked by specialist detectors trained on datasets such as MS-COCO \citep{mscoco}; substructure counting often requires open-vocabulary recognition and parent-object reasoning, making it a natural testbed for MLLM verifiers. We use this split to ask whether an automatic benchmark can reliably verify counting claims.

\section{Structural Counting Audit Suite}
\label{sec:audit-suite}

\paragraph{\longwang{Object}s and prompts.}
We include two structural-counting regimes. Object counting asks for the number of visible instances of a common object category. We choose three MS-COCO categories---cell phone, chair, and traffic light---so that detector-style baselines are meaningful for at least part of the suite \citep{mscoco}. Parent-bound substructure counting asks for the number of visible parts attached to a single main object. We use clover leaflets, dresser drawer fronts, power-strip sockets, and turbine blades. These \longwang{objects} cover several spatial layouts: radial arrangements, parallel or near-parallel rows, and repeated small parts.

\longwang{We provide three prompt templates for each object}. Each template contains a count variable: $K$ for object counts and $C$ for substructure counts. Within each generator, a \emph{cell} \longwang{corresponds to a combination of object, target count, and prompt template}. The full release contains 60 images per object-counting cell and 50 images per substructure-counting cell for each generator. Table~\ref{tab:audit-suite} summarizes the design.

\begin{table}[t]
\centering
\scriptsize
\setlength{\tabcolsep}{3pt}
\renewcommand{\arraystretch}{0.88}
\caption{Structural counting audit suite. Obj denotes instance-level object counting; Part denotes parent-bound substructure counting. Each \longwang{object} has three prompt templates.}
\label{tab:audit-suite}
\begin{tabular}{@{}l c c r r@{}}
\toprule
\longwang{Object} & Type & Targets & Imgs/cell & Imgs/gen. \\
\midrule
Cell phone & Obj & $K=2,3,4$ & 60 & 540 \\
Chair & Obj & $K=2,3,4$ & 60 & 540 \\
Traffic light & Obj & $K=2,3,4$ & 60 & 540 \\
Clover leaflets & Part & $C=3,4$ & 50 & 300 \\
Dresser fronts & Part & $C=4,5,6$ & 50 & 450 \\
Power-strip sockets & Part & $C=4,5,6$ & 50 & 450 \\
Turbine blades & Part & $C=3,4$ & 50 & 300 \\
\midrule
Total & -- & -- & -- & 3,120 \\
\bottomrule
\end{tabular}
\end{table}

\paragraph{Task selection and freezing.}
The task set is purposive rather than a random sample of visual capabilities. The recorded design history includes 1--3 manual template refinements to clarify the counting unit, followed by freezing the 57 structural cells before verifier sweeps. Verifier outputs were not used to choose tasks or tune prompts. The evaluation takes the first 20 image positions in each generator--cell without filtering by human or verifier outcomes.

\paragraph{Human audit.}
For each image, annotators record both a status label and, when the image is countable, a human-visible count. A \emph{valid} image has a clear counting scope and the requested target units can be counted reliably. An \emph{ambiguous} image contains relevant visual evidence but the count cannot be reliably determined because of blur, occlusion, cropping, merged units, malformed structure, or unclear boundaries. A
\emph{clear-fail} image is visibly unsuitable for the intended counting
task, for example because the target object is missing, the wrong object
is shown, multiple competing parent objects appear in a close-up
substructure task, or extra target objects invalidate the intended
counting scope.

For benchmark scoring, we collapse \emph{ambiguous} and \emph{clear-fail} into a single \emph{reject} class. This is because \longwang{neither case should} receive a successful structural-counting label under the benchmark definition. The human benchmark pass label is $h_i=\mathbf{1}[s_i^H=\valid \land c_i^H=t_i]$, where $t_i$ is the requested target count, $s_i^H$ is the human status, and $c_i^H$ is the human count when the image is valid.

The primary benchmark labels are produced by Human 1 (H1) using the same \longwang{object} card
and counting-scope instructions provided to the MLLM verifiers. With the same instructions, Human 2 (H2)
independently re-annotates a random broad subset and the full traffic-light
test set, and Human 3 (H3) independently labels the same images. H2 supports
inter-annotator reliability and disagreement sensitivity analyses, and H3 a reference-sensitivity check; the
released metadata includes all Human-1, Human-2 and Human-3 labels.

\paragraph{Reference provenance.}
H1, H2 and H3 are paper authors. H2's broad and traffic-light audits contain 640 context-specific records for 600 unique images, with 40 overlapping identities. Three overlapping images have differing context records; the canonical matched audit uses their traffic-light-context H2 labels. H3 labels these 600 images without access to prior labels, verifier outputs or generator identity, but with the requested target count visible, making this a label- and source-blind sensitivity check. Appendix~\ref{app:camera-ready-updates} reports this matched audit separately from the original context-specific analyses below.

\paragraph{Multi-annotator audit.}
To estimate the reliability of the human labels, we conducted an
independent second audit on a held-out sample of 280 images. We additionally audited all 360 images in the traffic-light test set to assess human-label reliability under a hard rejection-boundary setting. In this analysis we use the same two-class status label as the benchmark scoring rule:
\emph{valid} versus \emph{reject}, where reject collapses ambiguous and
clear-fail images.

Table~\ref{tab:double-human-audit} summarizes the results. On the
280-image audit sample, H1 and H2 reach 94.29\% exact agreement
on valid/reject status, with Cohen's $\kappa=0.636$. When both mark an image as valid, their exact count agreement is 98.39\%. The final benchmark pass/fail labels agree on 94.64\% of images, with $\kappa=0.844$. Thus, numeric count labels are highly stable once the counting scope is agreed upon. The main residual uncertainty is the valid/reject boundary.

The traffic-light stress slice has lower status agreement
(85.56\%, $\kappa=0.707$) and lower benchmark-success agreement
(86.39\%). This confirms that
traffic-light generations are a hard rejection-boundary case. We therefore report rejection behavior separately from valid-image count fidelity. The H3 labels support a three-annotator consensus sensitivity: retaining H1 outside the audited packet and excluding six unresolved IDs symmetrically leaves 2,274 images. Mean $\kappa$ over the current 20 configurations changes from $0.539$ to $0.563$, with paired difference $0.024$ and 95\% interval $[0.014,0.034]$ (Appendix~\ref{app:camera-ready-updates}).

\begin{table}[t]
\centering
\scriptsize
\setlength{\tabcolsep}{2pt}
\caption{
H1--H2 audit reliability. Reject collapses ambiguous and clear-fail
images, matching the benchmark scoring rule. Count agreement is computed
only on images marked valid by both annotators.
}
\label{tab:double-human-audit}
\begin{tabular}{lrrrrrrrr}
\toprule
Subset & $n$ & Stat. agr. & Stat. $\kappa$ & Valid F1 & Reject F1
& Count agr. & Pass agr. & Pass $\kappa$ \\
\midrule
Broad \longwang{object}--generator sample & 280 & 94.29\% & 0.636 & 96.88\% & 66.67\%
& 98.39\% & 94.64\% & 0.844 \\
Traffic-light test & 360 & 85.56\% & 0.707 & 87.32\% & 83.23\%
& 100.00\% & 86.39\% & 0.724 \\
\bottomrule
\end{tabular}
\end{table}

\paragraph{Evaluation split.}
The full audit suite contains 6,240 human-audited images across two generators. The main evaluation uses the first 20 independent seed-indexed generations from each cell as a held-out test split, yielding 1,140 images per generator and 2,280 images in total; MLLM few-shot support is selected outside these first-20 test positions. Detector support and evaluation splits are audited separately in Appendix~\ref{app:camera-ready-updates}. Unless otherwise stated, main-text results use this balanced subset. Main marginal intervals use 1,000 image-level percentile bootstrap resamples; paired comparison and reference-sensitivity intervals use 10,000 stratified resamples (Appendix~\ref{app:camera-ready-updates}). Appendix~\ref{app:dataset} reports full annotation statistics; for example, most \longwang{object}s have high valid rates, but traffic lights are a distribution-shift stress case, with valid rates of 87.78\% for Qwen-Image-2512 and 15.56\% for FLUX.2-dev after our revised traffic-light annotation pass. The artifact inventory includes generated images, prompts, target counts, human labels, verifier outputs, support-example indices, detector predictions, evaluation scripts, and metadata; Appendix~\ref{app:release-artifact} lists these components.

\section{Verifiers, Verifier-Use Protocols, and Metrics}
\label{sec:verifiers-protocols-metrics}

We separate three design choices that are often conflated: the verifier model $V$, the verifier-use protocol $\pi$, and the scoring rule $\sigma$. The same verifier can be used under different protocols, and the same per-image verifier output can be scored under different rules.

\paragraph{Verifier models.}
The main open-model audit includes Qwen3.5-4B/9B/27B~\citep{qwen35,qwen35_27b,qwen35_4b,qwen35_9b}, Gemma 3n E4B-it~\citep{gemma3n} and Gemma 4 E4B-it~\citep{gemma4}. We also run GPT-5.5~\citep{gpt55,gpt55api} and Claude Opus 4.7~\citep{claudeopus47} as targeted frontier closed-model sanity checks on clover and traffic-light \longwang{object}s to test whether more capable MLLMs eliminate the hardest failure modes. Detector-style baselines are natural for the three COCO object categories and for support-set-trained substructure detectors; we include compact detector results in Section~\ref{sec:results} and full detector protocol details in Appendix~\ref{app:detectors}.

\paragraph{Verifier-use protocols.}
The main MLLM protocols are summarized in Table~\ref{tab:protocols}. The count-only protocol asks the verifier to output a count and cannot reject an image. The reject/count protocol asks for both a validity judgment and a count; its scoring rule only passes images that are both predicted valid and counted as the requested target. We evaluate both zero-shot and few-shot variants. Few-shot support uses examples from the same \longwang{object} and prompt family; it is treated as a medium-support protocol. We also analyze TIFA-like target queries in which the verifier is shown a candidate count and answers a binary question such as ``are there exactly $k$ objects?'' Appendix~\ref{app:prompt} contains the full prompt template.

\begin{table}[t]
\centering
\scriptsize
\setlength{\tabcolsep}{3.5pt}
\caption{Verifier-use protocols and scoring rules.}
\label{tab:protocols}
\begin{tabular}{p{2.7cm} c c c p{5.5cm}}
\toprule
Protocol & Sees target? & \longwang{object} card? & Can reject? & Per-image output and scoring rule \\
\midrule
Count-only MLLM & no & yes & no & JSON integer count $\hat c_i$; auto pass iff $\hat c_i=t_i$. \\
\midrule
Reject/count MLLM & no & yes & yes & JSON validity decision $\hat s_i$ and count $\hat c_i$; auto pass iff $\hat s_i=\valid$ and $\hat c_i=t_i$. \\
\midrule
TIFA-like target query & yes & optional & no & Yes/no answer to ``exactly $k$?''; image-level diagnostics ask all candidate counts and require exactly one accepted answer. \\
\midrule
Detector-count & no & no & no & Auto pass iff detector count equals $t_i$. \\
\bottomrule
\end{tabular}
\end{table}

\paragraph{Metrics.}
We report six complementary metrics. The human benchmark score is $H=\frac{1}{N}\sum_i h_i$. The automatic benchmark score is $A=\frac{1}{N}\sum_i a_i$, where $a_i$ is the auto pass label induced by the chosen scoring rule. The aggregate score gap is $A-H$. Image-level agreement and Cohen's $\kappa$ compare $a_i$ and $h_i$. Valid-image count fidelity is $\Pr[\hat c_i=c_i^H\mid s_i^H=\valid]$, with rejection and nonnumeric outputs counted as failures on this full human-valid denominator. Count MAE and signed bias instead condition on numeric outputs and must be interpreted alongside numeric-output coverage. For reject/count protocols, we also report rejection precision, recall, and F1 against $\mathbf{1}[s_i^H\ne\valid]$. Finally, on human-valid images we report count MAE, signed bias, and the directional-error ratio $\rho_{\mathrm{bias}}=|\mathrm{bias}|/\mathrm{MAE}$ for $\mathrm{MAE}\neq 0$, where values near one mean that numeric errors are mostly one-directional rather than random. For two generators $g_1,g_2$, the comparison distortion $D=[A(g_1)-A(g_2)]-[H(g_1)-H(g_2)]=b(g_1)-b(g_2)$, with $b(g)=A(g)-H(g)$, measures how much the verifier changes the human difference between them.

\paragraph{Execution provenance.}
A verifier configuration includes rendered text/image order, image preprocessing, backend, precision, parser and output schema. The submitted Gemma matrix used author-side longest-edge-384 resizing. Some historical Qwen few-shot executions did not preserve the intended multimodal interleaving; Section~\ref{subsec:rq2} quantifies the effect. Appendix~\ref{app:camera-ready-updates} reports the current counting matrix, its uncertainty analyses and the spatial and support-set reruns.

\section{Results}
\label{sec:results}
\noindent\textit{Execution scope.} Open-MLLM results below use the original-input matrix: 18,240 new Gemma judgments and 27,360 native-interleaved Qwen judgments. Model-size effects remain confounded with execution differences.

\subsection{RQ1: Does aggregate score matching imply reliable verification?}
\label{subsec:rq1}

\begin{figure}[t]
\centering
\includegraphics[width=0.75\linewidth]{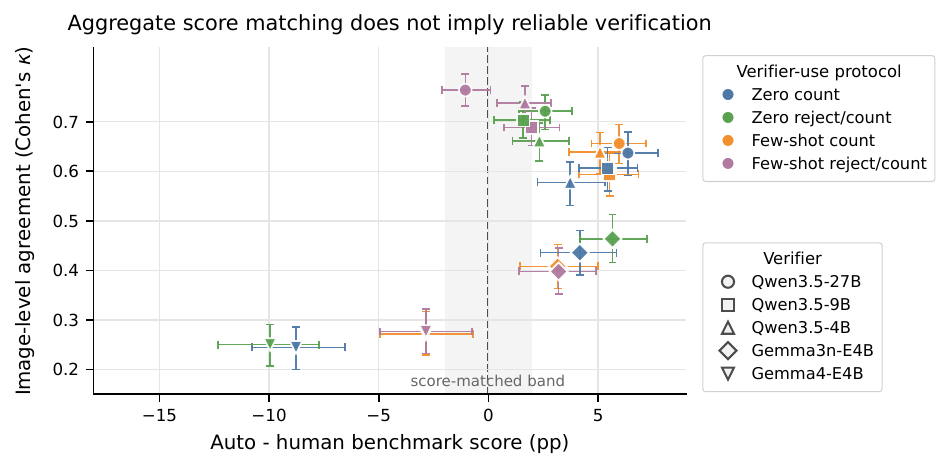}
\caption{Aggregate score gap and image-level agreement for the current verifier--protocol pairs. The shaded band marks $\pm2\pp$ around score parity as a visual reference. Error bars are marginal 95\% image-bootstrap intervals.}
\label{fig:score-vs-kappa}
\end{figure}

\textbf{No.} Aggregate score matching can coexist with image-level disagreement. Figure~\ref{fig:score-vs-kappa} plots the automatic-minus-human benchmark score gap against Cohen's $\kappa$ for MLLM verifier-use protocols. Gemma4-E4B few-shot reject/count and Qwen3.5-27B few-shot reject/count both have score gaps within $3\pp$ of zero ($-2.85\pp$ and $-1.05\pp$). Yet their image-level reliability differs substantially: $\kappa=0.277$ versus $0.764$, and valid-image count fidelity $78.53\%$ versus $93.24\%$. A small aggregate score gap is therefore insufficient evidence that an automatic verifier selects the same passing images as humans.

Detector baselines (Table~\ref{tab:main-detector-protocols}) show the same aggregate-versus-image-level distinction. A COCO-pretrained YOLO11s detector \citep{yolo11s} on the Qwen-Image object-count subset improves substantially under oracle object-wise threshold selection, moving from $A-H=-13.58\pp$, $\kappa=0.294$ to $A-H=-3.14\pp$, $\kappa=0.519$. However, support-trained YOLO11s on the balanced object split (cell phones, chairs and traffic lights) has a small aggregate gap ($A-H=+4.63\pp$) only because under-scoring on phones and chairs ($-17.50$ and $-10.83\pp$) offsets over-scoring on traffic lights ($+42.22\pp$; Appendix~\ref{app:detectors}), and its pass/fail agreement is barely above chance ($\kappa=0.016$). Thus, for detector protocols too, matching the aggregate score can still hide severe image-level disagreement.

\textbf{Matching the pooled score also does not preserve generator differences.} Humans score Qwen-Image-2512 $25.44\pp$ above FLUX.2-dev. All 20 MLLM configurations shrink this difference: the comparison distortion $D$ ranges from $-7.54\pp$ to $-25.00\pp$, and every fixed-cell 95\% interval excludes zero (Table~\ref{tab:distortion-current}). Gemma4-E4B few-shot reject/count, whose pooled gap is only $-2.85\pp$, nearly erases the generator difference ($0.44\pp$), and even the highest-agreement configuration, Qwen3.5-27B few-shot reject/count, reduces it to $16.32\pp$ ($D=-9.12\pp$). The pooled gap, image-level agreement and $D$ are therefore distinct dimensions of verifier reliability.

\begin{table}[t]
\centering
\scriptsize
\setlength{\tabcolsep}{4pt}
\caption{Detector baselines also depend on verifier-use protocol. COCO rows use the pretrained YOLO11s detector on the Qwen-Image prediction-covered subset; the task-wise thresholds are an oracle diagnostic selected on that same subset. Support-trained rows use the reported balanced splits; the substructure row excludes 57 test images that overlap its support set (1,200-image result in Appendix~\ref{app:camera-ready-updates}).}
\label{tab:main-detector-protocols}
\begin{tabular}{llrrrr}
\toprule
Detector protocol & Scope & $n$ & $A-H$ pp & Count valid \% & $\kappa$ \\
\midrule
COCO YOLO11s, fixed $.25$ & Qwen obj pred & 1561 & -13.58 & 85.47 & 0.294 \\
COCO YOLO11s, oracle \longwang{object} threshold & Qwen obj pred & 1561 & -3.14 & 95.57 & 0.519 \\
YOLO11s object, support-trained & MLLM obj split & 1080 & +4.63 & 87.18 & 0.016 \\
YOLO11s substructure, support-trained & MLLM part split & 1143 & -0.79 & 94.84 & 0.862 \\
\bottomrule
\end{tabular}
\end{table}

\subsection{RQ2: Which verifier-use choices change the measurement?}
\label{subsec:rq2}

\begin{figure}[t]
\centering
\begin{subfigure}{0.60\linewidth}
    \centering
    \includegraphics[width=\linewidth]{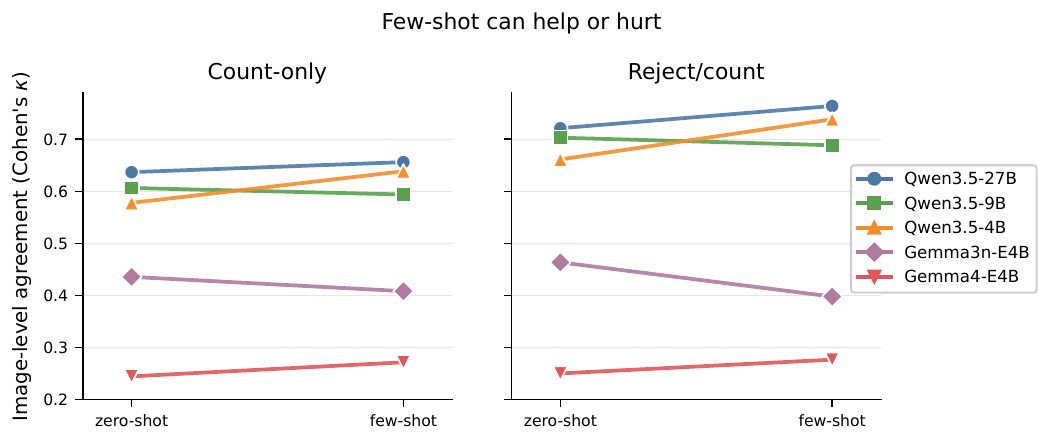}
    \caption{Few-shot support can help or hurt.}
\end{subfigure}\hfill
\begin{subfigure}{0.35\linewidth}
    \centering
    \includegraphics[width=\linewidth]{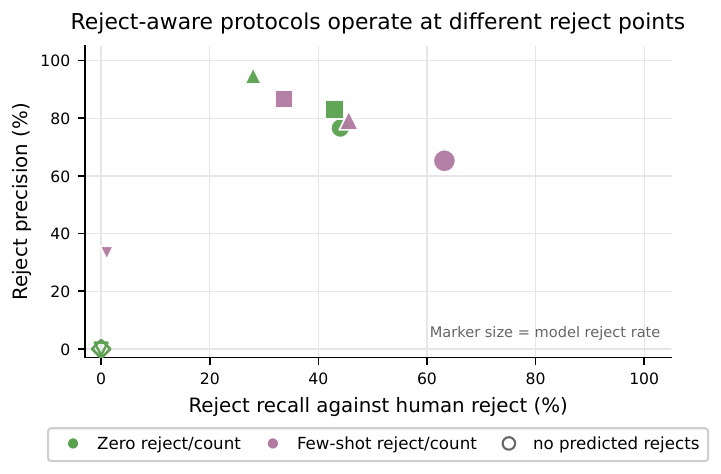}
    \caption{Reject/count protocols operate at different reject points.}
\end{subfigure}
\caption{Verifier-use protocols are measurement variables. Support and rejection affordances have model-dependent effects. In panel (b), open markers at the origin denote no predicted rejects, for which precision is undefined; marker size encodes rejection rate.}
\label{fig:fewshot-reject}
\end{figure}

\textbf{Verifier-use protocols change both aggregate scores and image-level agreement.} Qwen3.5-27B few-shot reject/count has the highest agreement in the current matrix: relative to few-shot count-only, $A-H$ changes from $+5.96\pp$ to $-1.05\pp$ and $\kappa$ rises from $0.656$ to $0.764$. Its full-denominator count fidelity falls from $96.50\%$ to $93.24\%$, illustrating a trade-off between rejection and count recovery. Support effects also vary: moving from zero-shot to few-shot reject/count raises Qwen3.5-4B $\kappa$ from $0.661$ to $0.738$, but lowers Gemma3n-E4B $\kappa$ from $0.464$ to $0.398$ and count fidelity from $90.27\%$ to $87.73\%$. Figure~\ref{fig:fewshot-reject} summarizes these configuration-dependent effects.

Support effects also depend on how the multimodal prompt is rendered. On a 228-image Qwen3.5-9B reject/count subset with fixed weights, images, support examples and prompts, a renderer that places the support images ahead of the text, away from their labels, lowers $\kappa$ from $0.691$ zero-shot to $0.334$ few-shot ($\Delta\kappa=-0.357$, 95\% interval $[-0.513,-0.203]$). Under three renderers that keep each support image next to its label, few-shot support changes $\kappa$ by only $+0.039$ to $+0.065$, with intervals including zero (Table~\ref{tab:renderer-ablation}).

The reject/count protocol is especially important under noisy generator distributions, but its effect is model-dependent. Different MLLMs choose different reject operating points, as shown in Figure~\ref{fig:fewshot-reject}b. A useful reject-aware verifier must be evaluated by full-denominator benchmark score, image-level agreement, and rejection precision/recall, not only by valid-image count accuracy.

TIFA-like target queries expose a complementary failure mode. On 1,140 human-valid Qwen-Image images, the Qwen3.5-27B \longwang{object}-guided target-query run answers 5,819 usable yes/no queries with 97.11\% accuracy, but only 90.44\% of images accept exactly the human count and no other; 2.63\% accept none, 1.75\% several, and 5.18\% a single wrong count. The failure is \longwang{object}-local: chairs and traffic lights reach 100\% image-level accuracy, while dresser fronts reach only 63.33\%. High query-level accuracy therefore does not imply a correct image-level count.

Detector reliability likewise depends on the target unit: the support-trained substructure detector reaches $\kappa=0.862$, whereas the support-trained object detector barely exceeds chance agreement despite high valid-image count accuracy (Table~\ref{tab:main-detector-protocols}). Detectors suit visually regular, boxable targets in clean, low-reject settings, but are weaker as standalone verifiers when the benchmark requires semantic validity decisions, background-object exclusion, or rejection of malformed generations.

\subsection{RQ3: Where do failures localize, and do they affect generator conclusions?}
\label{subsec:rq3}

\begin{figure}[t]
\centering
\includegraphics[width=0.9\linewidth]{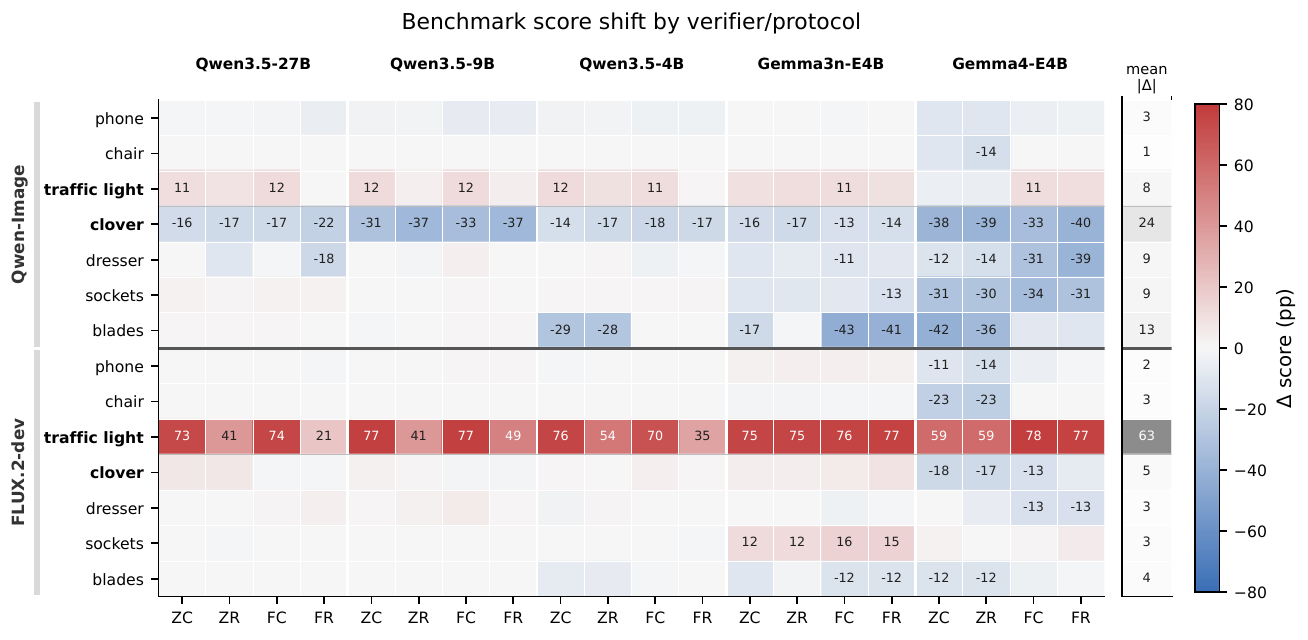}
\caption{\longwang{object}-local benchmark score shifts under verifier-use choices. Cells show automatic minus human benchmark score in percentage points. Column labels abbreviate protocols: Z/F denote zero-/few-shot, and C/R denote count-only/reject-count. Large positive gaps often correspond to false passes on human-rejected images; large negative gaps often correspond to substructure counting failures.}
\label{fig:task-heatmap}
\end{figure}

\textbf{Failures are highly task-local and directional}. Figure~\ref{fig:task-heatmap} shows the per-task automatic--human score gap for each generator, verifier, and protocol. Large distortions arise from plausible counts on human-rejected images and errors in fine-grained substructures. On FLUX.2-dev traffic lights, Gemma3n few-shot reject/count over-scores humans by $76.67\pp$, Qwen3.5-9B zero-shot count by $77.22\pp$, and Qwen3.5-27B few-shot count by $74.44\pp$. On Qwen-Image-2512 clover, few-shot reject/count under-scores humans by $40.00\pp$ for Gemma4-E4B, $36.67\pp$ for Qwen3.5-9B, and $22.50\pp$ for Qwen3.5-27B. Overall benchmark averages hide these local failures.

These local failures can also change \longwang{object}-level generator comparisons. Using
the human benchmark score as the reference ordering between Qwen-Image-2512 and
FLUX.2-dev, the all-image automatic benchmark score flips 11 of 140
verifier--protocol--\longwang{object} comparisons. The effect
is stronger for diagnostics restricted to human-valid images: valid-only target pass
flips 20 of 140 comparisons, and valid-image count fidelity flips 56 of 140.
These diagnostics change the estimand; their reversals describe scoring-rule sensitivity.
The flips concentrate in substructure \longwang{object}s such as clover, dresser fronts, and
power-strip sockets. Appendix~\ref{app:ranking-flips} visualizes the \longwang{object}-level
flip pattern.

\textbf{With a third generator, verifier errors reverse a human ranking.} We also generated the 57 structural cells with Z-Image~\citep{zimage} and scored 1,140 images per generator under a common Qwen3.5-4B original-input configuration; Z-Image has Human-1 labels only. Humans order Qwen-Image ($92.02\%$) above Z-Image ($71.32\%$) above FLUX.2-dev ($66.58\%$), but all four protocols place FLUX.2-dev above Z-Image: the automatic Z-Image-minus-FLUX difference is between $-3.07$ and $-6.40\pp$ (all fixed-cell intervals below zero) against a human difference of $+4.74\pp$, so $D$ lies between $-7.81$ and $-11.14\pp$, with every interval excluding zero (Appendix~\ref{app:camera-ready-updates}). Z-Image itself is scored accurately ($|A-H|\le2.63\pp$, $\kappa\ge0.778$); the reversal comes from over-scoring FLUX.2-dev by $5.44$--$11.23\pp$. It holds on the fixed suite, where the human difference has fixed-cell interval $[1.93,7.54]\pp$. Across a broader cell population the human ordering itself is uncertain (hierarchical interval $[-4.39,14.12]\pp$), and removing traffic lights changes both orderings.

\begin{figure}[t]
\centering
\includegraphics[width=0.9\linewidth]{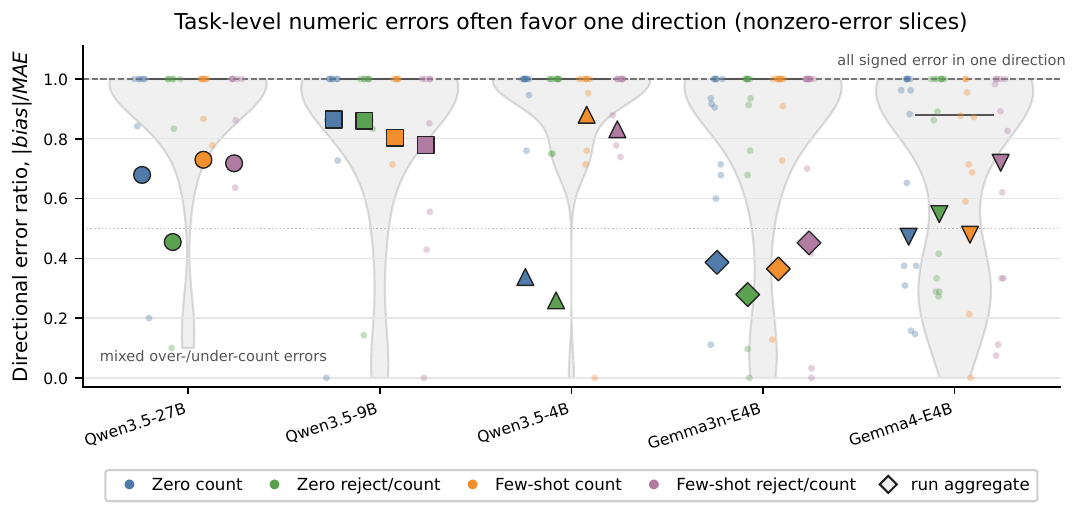}
\caption{Directional counting errors across all seven tasks. Distributions use generator--task slices with nonzero MAE, conditioning numeric errors on human-valid images with numeric predictions; full-run markers use the same conditional metric.}
\label{fig:bias-ratio}
\end{figure}

Count errors are also directional. Figure~\ref{fig:bias-ratio} plots $\rho_{\mathrm{bias}}$ across generator--task slices after removing zero-MAE slices. On the full seven-task population, Qwen3.5-27B zero-shot count has conditional numeric MAE $0.0388$ and signed bias $+0.0264$, giving $\rho_{\mathrm{bias}}=0.679$. The corresponding ratio is $0.718$ for Qwen3.5-27B few-shot reject/count and $0.864$ for Qwen3.5-9B zero-shot count, both with positive signed bias. Table~\ref{tab:current-numeric-errors} reports numeric-output coverage. Directional bias adds information beyond aggregate pass rates when generator outputs shift toward tasks with one-sided count errors.

Targeted frontier sanity checks lead to the same conclusion. On a 600-image clover-plus-traffic slice, GPT-5.5 is closer to the human all-image benchmark rate than Claude Opus 4.7 ($A-H=-4.33\pp$ versus $-12.50\pp$; $\kappa=0.522$ versus $0.432$). GPT-5.5 has strong traffic-light valid-count fidelity ($94.27\%$) but still over-scores humans by $10.56\pp$, while its clover count fidelity falls to $59.92\%$ with signed bias $+0.543$ leaflets. Claude avoids the traffic-light over-score ($A-H=-5.00\pp$), but has lower valid-count fidelity on both traffic lights ($74.48\%$) and clover ($56.96\%$). A frontier MLLM is therefore a useful sanity check, not a substitute for a factorized verifier reliability profile.

\section{Discussion and Limitations}
\label{sec:discussion-limitations}
\label{sec:selection-card}

The main empirical conclusion is that structural-counting scores depend on the verifier, verifier-use protocol, scoring rule, and generator distribution. We recommend that T2I benchmark authors publish a compact \emph{verifier selection card} alongside every automatic score, reporting at least the verifier/version, query protocol, support examples or thresholds, scoring rule, human-audit split, aggregate score gap, image-level $\kappa$, the distortion $D$ of each generator comparison the benchmark is meant to resolve, valid-image count fidelity, rejection metrics when applicable, and \longwang{object}-level signed bias. When the interval for $D$ exceeds the comparison error the benchmark can tolerate, the comparison should be reported as unresolved. Appendix~\ref{app:selection-card} gives a template.

For structural counting, verifier suitability should be tied to the target visual unit and rejection policy, not only to the broad label ``object'' or ``part'' counting. Clean common-object counting can use count-only verifiers as a low-cost first pass, but they must be stress-tested against human-rejected images; parent-bound substructures require treating support examples, rejection policy, and \longwang{object}-local shifts as part of the measurement.

\paragraph{Limitations.}
\label{sec:limitations}

Our audit is scoped to structural counting, three prompt templates per object, two generators, and a finite set of verifier-use protocols. The third-generator comparison uses one verifier configuration and Human-1 labels only, and the spatial and real-image-counting extensions in Appendix~\ref{app:camera-ready-updates} test specific boundaries rather than general compositional evaluation. Human labels target counting-scope reliability rather than general prompt satisfaction, and frontier closed-model checks are not fully reproducible. The main measurement lesson is that automatic benchmark scores should be reported with the verifier, protocol, scoring rule, and reliability diagnostics that produced them.

\input{ack}

\bibliographystyle{plainnat}
\bibliography{refs}

\appendix

\section{Dataset Details and Human Annotation Statistics}
\label{app:dataset}

Table~\ref{tab:full-validity} reports the revised full-suite human status distribution. The most important distribution-shift case is FLUX.2-dev traffic lights: many generations contain multiple light heads on a single pole or other clear-fail structures, so the valid rate is only 15.56\%. This is useful for auditing whether a verifier can distinguish clean counting from benchmark-invalid generations.

\paragraph{Evaluation and audit sampling.}
The balanced MLLM evaluation split uses the first 20 seed-indexed generations
from every generator--\longwang{object}--prompt--target-count cell; no image is filtered by
human validity before verifier evaluation. MLLM few-shot support is selected
outside the first-20 evaluation positions. This exclusion applies to MLLM
support; detector training/support splits have a separate
overlap audit in Appendix~\ref{app:camera-ready-updates}. The
280-image broad H2 audit sample is stratified by generator and \longwang{object}:
we sample 20 images from each of the $2\times7$ generator--\longwang{object} strata within
the balanced first-20 split. The traffic-light stress audit is the complete
traffic-light portion of the same split, i.e., $2$ generators $\times$ $3$
prompt templates $\times$ $3$ target counts $\times$ $20$ images, for 360
images.

\begin{table}[h]
\centering
\scriptsize
\setlength{\tabcolsep}{3pt}
\caption{Full human-audited status counts by generator and \longwang{object} after the revised traffic-light annotation pass.}
\label{tab:full-validity}
\begin{tabular}{llrrrr}
\toprule
Generator & \longwang{Object} & Valid & Ambiguous & Clear-fail & Valid rate \\
\midrule
Qwen-Image-2512 & cell phone & 536 & 4 & 0 & 99.26\% \\
Qwen-Image-2512 & chair & 539 & 1 & 0 & 99.81\% \\
Qwen-Image-2512 & clover leaflets & 288 & 12 & 0 & 96.00\% \\
Qwen-Image-2512 & dresser fronts & 441 & 9 & 0 & 98.00\% \\
Qwen-Image-2512 & power-strip sockets & 431 & 4 & 15 & 95.78\% \\
Qwen-Image-2512 & traffic lights & 474 & 16 & 50 & 87.78\% \\
Qwen-Image-2512 & turbine blades & 299 & 1 & 0 & 99.67\% \\
\midrule
FLUX.2-dev & cell phone & 532 & 7 & 1 & 98.52\% \\
FLUX.2-dev & chair & 539 & 1 & 0 & 99.81\% \\
FLUX.2-dev & clover leaflets & 298 & 2 & 0 & 99.33\% \\
FLUX.2-dev & dresser fronts & 444 & 6 & 0 & 98.67\% \\
FLUX.2-dev & power-strip sockets & 443 & 7 & 0 & 98.44\% \\
FLUX.2-dev & traffic lights & 84 & 2 & 454 & 15.56\% \\
FLUX.2-dev & turbine blades & 297 & 3 & 0 & 99.00\% \\
\bottomrule
\end{tabular}
\end{table}

\paragraph{Annotation categories.}
\textbf{Valid} means the target units are visible and have a well-defined counting scope. \textbf{Ambiguous} means the image is plausibly relevant, but the target count is not reliably determined because of blur, occlusion, cropping, merged units, malformed structure, or unclear boundaries. \textbf{Clear-fail} means the image should not be used as a structural counting instance under the intended \longwang{object} because the target is absent, the wrong object is shown, the parent object is duplicated in a way that breaks the substructure scope, or prompt-violating extra targets make the intended count undefined. Human benchmark pass labels require both valid status and exact count match.

\paragraph{Traffic-light boundary rule.}
For traffic-light object counting, the validity label follows the counting
scope rather than every scene-level prompt constraint. The counted units are
the clear, upright traffic lights of the prompted style in the main counting scene,
each one pole with one vertical signal head.
Background signals that are clearly a different type, such as distant
horizontal road-intersection signal bars, are ignored and do not by themselves
make the image invalid. A pole with several signal heads, or a clearly visible
background light of the same style that could be counted as another unit, makes
the image reject because the counting scope is violated, even when every light
is easy to see. Very
blurred or unidentifiable background marks are ignored rather than counted.

\section{Released Artifact Inventory}
\label{app:release-artifact}

The artifact is organized as a data release (\url{https://huggingface.co/datasets/neurips2026-ed443/audit-release}) and a code release (\url{https://github.com/AzusaCat1111/verifier-choice-audit}). The dataset repository includes Croissant metadata with Responsible-AI fields. The
data release contains generated images, human annotations, prompt metadata,
verifier outputs, detector outputs, and analysis tables. The code release
contains the scripts used for scoring, bootstrap confidence intervals,
detector evaluation, aggregation, and table generation, with the data release mounted
as \texttt{./dataset}. Table~\ref{tab:released-components} summarizes the
released components.

\begin{table}[h]
\centering
\small
\setlength{\tabcolsep}{5pt}
\caption{Released audit artifact components.}
\label{tab:released-components}
\begin{tabular}{p{0.29\linewidth}p{0.61\linewidth}}
\toprule
Released component & Contents \\
\midrule
Generated images & All images by generator, \longwang{object}, prompt template, target count, and seed \\
Human annotations & Human-1 full labels, Human-2 subset plus full traffic-light labels, and Human-3 labels for the same 600 images \\
Verifier outputs & Raw outputs, parsed JSON/counts, and parser status \\
Protocol metadata & Prompts, \longwang{object} cards, support indices, and target counts \\
Evaluation code & Scoring, bootstrap confidence intervals, aggregation, and table generation \\
Detector artifacts & Detector predictions, thresholds, and support/training metadata \\
\bottomrule
\end{tabular}
\end{table}

\section{Compute Resources}
\label{app:compute}

The released scoring, bootstrap, aggregation, and evaluation scripts are
CPU-bound and run on a standard workstation; the largest local data requirement
is the image/CSV artifact storage. Local detector experiments used a workstation
with three NVIDIA RTX 3090 GPUs (24GB each); YOLO training/evaluation used a
single GPU per run and completed within minutes to tens of minutes for each
object or substructure setting. Execution records for the historical open-MLLM runs, including GPU-hours, are incomplete, and the
four-GPU RTX PRO 6000 setup stated at submission is not confirmed for every run.
The current original-input campaign uses two local NVIDIA RTX A6000 GPUs
(48GB each), one per small-model lane and both for Qwen3.5-27B. Per-run
provenance distinguishes these executions from the historical endpoints. The main open-MLLM matrix evaluates 20
verifier--protocol rows on 2,280 images per row, while the frontier sanity checks
evaluate 600 images per closed model through hosted APIs. Wall-clock time for
hosted or locally served inference depends on endpoint throughput and rate
limits. We release run configurations, raw outputs, parsed outputs, and
evaluation scripts so the reported scores can be re-computed without rerunning
model inference. Preliminary and failed prompt/API experiments used additional
inference calls but are not needed to reproduce the reported tables.

\section{Generator-Ranking Flip Sensitivity}
\label{app:ranking-flips}

\begin{figure}[h]
\centering
\includegraphics[width=0.92\linewidth]{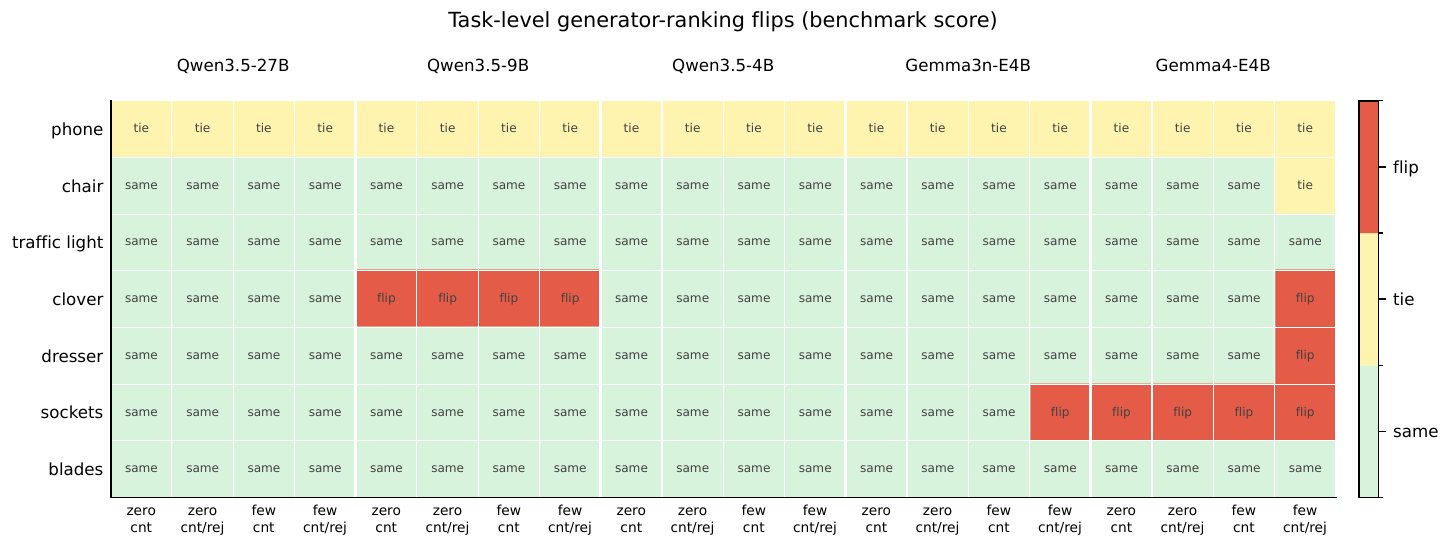}
\caption{\longwang{object}-level Qwen-Image versus FLUX.2-dev generator-ranking flips under
the all-image automatic benchmark score. Green cells preserve the human
generator ordering, yellow cells are ties under either human or automatic score,
and red cells flip the human ordering.}
\label{fig:ranking-flips}
\end{figure}

For each \longwang{object}, we compare the Qwen-Image-2512 versus FLUX.2-dev ordering under
the human benchmark score and under each automatic verifier--protocol score. A
ranking flip means that the automatic score orders the two generators opposite
to the human score on the same \longwang{object}. Figure~\ref{fig:ranking-flips} shows the
all-image benchmark-score flips. The main-text summary counts 11 flips among
140 verifier--protocol--\longwang{object} comparisons, with flips concentrated in
substructure \longwang{object}s.

\section{MLLM Prompt and Parsing}
\label{app:prompt}

The reject/count protocol uses a JSON schema so that output compliance can be measured separately from semantic correctness. A representative prompt is shown below; \longwang{object}-specific fields define the target units, ignored visual elements, and annotation boundary.

\paragraph{Few-shot support construction.}
Few-shot support examples are selected from images outside the balanced
evaluation split. For a query image from prompt family $p_j$, support is drawn
from the same generator, \longwang{object}, and prompt family $p_j$, so the support examples
match the query distribution without reusing target test images. Within that
bank, we take one human-valid example from each target-count cell in ascending
target order, preferring images whose human count equals the cell target, until
the support set holds three examples; families with fewer cells draw further
valid examples from the same cells.
For reject/count protocols, if a human-rejected example is available and the
protocol allows reject output, one of the three slots holds a reject example to expose the
valid/reject boundary. The support set is therefore fixed per generator--\longwang{object}--prompt
family and shared by every query in that family, so it does not depend on the query's target count.
One FLUX.2-dev traffic-light family has a single eligible valid example, so its prompts use that
example alone, plus one reject example under reject/count. The final support order is randomly permuted with a
fixed per-family seed, making the prompt deterministic while avoiding a fixed count-order
or label-order cue.

\begingroup\small
\begin{verbatim}
System:
You are a visual counting verifier for a generated-image benchmark.
Always return exactly one valid JSON object and no extra text.
Do not write step-by-step reasoning in the final answer.

User:
You are evaluating a generated image for a structural counting
   benchmark.
This is a reject-or-count protocol. First decide whether the final
   query
image is valid for this counting task. If it is valid, return the
   visual
count. If it is not reliably countable, reject it.

Count only the requested visible target units.
Target units to count: {target_units}
Do not count / ignore: {ignore}
Human annotation protocol for this task:
{annotation_protocol}

You will first see labeled examples from the same prompt family.
The examples are provided only to clarify the counting scope, the
   human
annotation protocol, the reject boundary, and the JSON format.
Do not copy the first example, the last example, the majority example,
or any typical count.

Validity labels:
- "valid": the image has a clear counting scope and the target units
   can be
  counted reliably.
- "reject": the image should not be counted because the target is
   missing,
  the wrong object is shown, the image violates the human annotation
   protocol,
  or the count is not reliable because of blur, occlusion, cropping,
   merged
  units, malformed structure, unclear boundaries, or unclear counting
   scope.

Rules:
1. Count only visible target units in the final query image.
2. Use the examples only to understand what should be counted and when
   to reject.
3. Do not copy any example count or validity label.
4. Do not infer the count from common sense, typical object shapes, or
   likely defaults.
5. Do not use any desired or requested count; no target count is
   provided.
6. If the final query image is valid, return an integer count.
7. If the final query image should be rejected, set count to null.

Return exactly one valid JSON object for the final query image:
{"validity": "valid" | "reject", "count": <integer or null>}
\end{verbatim}
\endgroup

\section{Overall MLLM Results}
\label{app:overall-results}

Table~\ref{tab:overall-results} gives a compact version of the overall open-MLLM results on the balanced evaluation split. Appendix~\ref{app:release-artifact} lists the released images, annotations, verifier outputs, protocol metadata, detector artifacts, and evaluation code.

\input{tables/app_overall_current.tex}
\input{tables/app_conditional_errors_current.tex}

\section{Target-Query / TIFA-like Counting Results}
\label{app:tifa-like}

The target-query protocol asks the verifier whether the image contains exactly a proposed count. Query-level binary accuracy can be high even when the induced image-level count is unreliable. The run uses Qwen-Image-2512 images and takes the first 20 human-valid images in each of the 57 cells, so it contains no human-rejected images. Each image is queried at every count from one below the smallest to one above the largest human count observed for its \longwang{object}, and the expected answer is yes only at the image's human count. On 5,819 usable yes/no query rows out of 5,820 target-query prompts, the Qwen3.5-27B \longwang{object}-guided target-query run has 97.11\% accuracy, 95.25\% balanced accuracy, and 95.40\% macro-F1. However, after aggregating over candidate counts per image, only 90.44\% of 1,140 images accept exactly the human count and no other. The remaining images accept no count (2.63\%), several counts (1.75\%) or a single wrong count (5.18\%; Table~\ref{tab:tifa-like-image}). The \longwang{object}-level image accuracy ranges from 100.00\% on chairs and traffic lights to 63.33\% on dresser fronts; clover errors are mostly single wrong counts (24.17\% of clover images), whereas dresser fronts also often accept no count (13.89\%). Because each image aggregates several queries, a single wrong answer is enough to make the image-level count incorrect. We therefore treat target-query accuracy as an insufficient reliability indicator unless accompanied by image-level aggregation over all candidate counts.

\begin{table}[h]
\centering
\small
\setlength{\tabcolsep}{4pt}
\caption{Image-level aggregation for the TIFA-like target-query counting protocol on 1,140 human-valid Qwen-Image images. The four outcome columns sum to 100\%: exactly the human count accepted (image acc.), no count accepted, several counts accepted, and a single wrong count accepted.}
\label{tab:tifa-like-image}
\begin{tabular}{lrrrrr}
\toprule
\longwang{Object} & Images & Image acc. & 0-yes rate & $>1$-yes rate & Wrong 1-yes rate \\
\midrule
Cell phone & 180 & 97.78\% & 2.22\% & 0.00\% & 0.00\% \\
Chair & 180 & 100.00\% & 0.00\% & 0.00\% & 0.00\% \\
Clover leaflets & 120 & 70.83\% & 0.83\% & 4.17\% & 24.17\% \\
Dresser fronts & 180 & 63.33\% & 13.89\% & 6.67\% & 16.11\% \\
Power-strip sockets & 180 & 98.33\% & 0.00\% & 1.11\% & 0.56\% \\
Traffic lights & 180 & 100.00\% & 0.00\% & 0.00\% & 0.00\% \\
Turbine blades & 120 & 99.17\% & 0.00\% & 0.83\% & 0.00\% \\
\midrule
Overall & 1140 & 90.44\% & 2.63\% & 1.75\% & 5.18\% \\
\bottomrule
\end{tabular}
\end{table}

\section{Targeted Frontier Closed-Model Sanity Checks}
\label{app:gpt55}

Table~\ref{tab:gpt55} reports targeted frontier closed-model sanity checks on clover and traffic lights. They test whether stronger closed MLLMs remove the most important failure modes. They do not: different frontier models shift the operating point, but clover counting and traffic-light rejection remain \longwang{object}-sensitive.

\begin{table}[h]
\centering
\scriptsize
\setlength{\tabcolsep}{3.0pt}
\caption{Frontier closed-model few-shot reject/count sanity checks on two hard \longwang{object}s.}
\label{tab:gpt55}
\begin{tabular}{llrrrrrrr}
\toprule
Model & \longwang{Object} & $n$ & H all \% & A all \% & $A-H$ pp & Count valid \% & Bias & $\kappa$ \\
\midrule
GPT-5.5 & Clover & 240 & 79.58 & 52.92 & -26.67 & 59.92 & +0.543 & 0.344 \\
Claude Opus 4.7 & Clover & 240 & 79.58 & 55.83 & -23.75 & 56.96 & +0.414 & 0.203 \\
GPT-5.5 & Traffic & 360 & 53.33 & 63.89 & +10.56 & 94.27 & 0.000 & 0.660 \\
Claude Opus 4.7 & Traffic & 360 & 53.33 & 48.33 & -5.00 & 74.48 & -0.070 & 0.557 \\
\midrule
GPT-5.5 & Overall & 600 & 63.83 & 59.50 & -4.33 & 75.29 & +0.300 & 0.522 \\
Claude Opus 4.7 & Overall & 600 & 63.83 & 51.33 & -12.50 & 64.80 & +0.212 & 0.432 \\
\bottomrule
\end{tabular}
\end{table}

\section{Detector Protocol Placement}
\label{app:detectors}

Specialist detectors provide a useful contrast to MLLM verifiers when the target units have labels that a detector can learn or already knows. For COCO object categories, an off-the-shelf detector can be used with a fixed category label and confidence threshold. For substructure \longwang{object}s, a detector requires support data and should therefore be reported as a medium-support verifier-use protocol. We recommend reporting detector results separately for same-generator and cross-generator evaluation because support images sampled from one generator can encode generator-specific artifacts. The detector selection card should include detector family and version, pretrained dataset, support-set size, annotation time, threshold search procedure, non-maximum-suppression settings, same-generator score, cross-generator score, and failure cases.

Table~\ref{tab:detector-baselines} reports the detector-count baselines using
the same benchmark pass/fail scoring rule as the MLLM verifiers. The first two
rows are the earlier zero-project-label COCO detector runs on the Qwen-Image
object-count prediction subset, re-scored against the current annotation JSON
files. The original script only emitted predictions for images that were valid
under the earlier annotation pass; after the revised traffic-light labels, the
prediction-covered subset misses 35 currently valid traffic-light images. The second COCO row uses task-wise thresholds ($0.50/0.25/0.75$ for phones/chairs/traffic lights) that maximize exact-count accuracy post hoc on the same 1,561 prediction-covered evaluation images, so it is an oracle diagnostic. The remaining rows use
support-trained YOLO11s detectors on the MLLM evaluation split. These
results are included as detector protocol reference points. They also illustrate why detector protocols should
be reported factorially: the substructure detector closely tracks human
pass/fail labels, while the object detector can still have low image-level
agreement when the generator produces validity-shifted traffic-light images.

\begin{table}[h]
\centering
\scriptsize
\setlength{\tabcolsep}{4pt}
\caption{Detector-count baselines. $A-H$ and $\kappa$ use image-level
benchmark pass/fail labels. The COCO YOLO11s rows are zero-project-label
pretrained detector diagnostics on the prediction-covered Qwen-Image object
subset, re-scored with current annotations; the support-trained rows use the
balanced MLLM evaluation split, and the object-detector task rows pool both generators. Substructure rows use all 1,200 test images, including the 57 that overlap the support set.}
\label{tab:detector-baselines}
\begin{tabular}{llrrrrrr}
\toprule
Detector protocol & Split & $n$ & H all \% & A all \% & $A-H$ pp & Count valid \% & $\kappa$ \\
\midrule
COCO YOLO11s, fixed $.25$ & Qwen obj pred & 1561 & 96.03 & 82.45 & -13.58 & 85.47 & 0.294 \\
COCO YOLO11s, \longwang{object} threshold & Qwen obj pred & 1561 & 96.03 & 92.89 & -3.14 & 95.57 & 0.519 \\
\midrule
Object detector & Overall & 1080 & 82.69 & 87.31 & +4.63 & 87.18 & 0.016 \\
Object detector & Qwen-Image & 540 & 94.44 & 92.22 & -2.22 & 92.46 & 0.109 \\
Object detector & FLUX.2-dev & 540 & 70.93 & 82.41 & +11.48 & 80.15 & -0.077 \\
Object detector & Cell phones & 360 & 95.00 & 77.50 & -17.50 & 79.34 & 0.175 \\
Object detector & Chairs & 360 & 99.72 & 88.89 & -10.83 & 88.89 & 0.044 \\
Object detector & Traffic lights & 360 & 53.33 & 95.56 & +42.22 & 98.44 & 0.065 \\
\midrule
Substructure detector & Overall & 1200 & 76.25 & 75.50 & -0.75 & 94.92 & 0.861 \\
Substructure detector & Qwen-Image & 600 & 89.83 & 88.50 & -1.33 & 96.42 & 0.759 \\
Substructure detector & FLUX.2-dev & 600 & 62.67 & 62.50 & -0.17 & 93.45 & 0.883 \\
\bottomrule
\end{tabular}
\end{table}

\clearpage
\section{Verifier Selection Card Template}
\label{app:selection-card}

\begin{table}[h]
\centering
\small
\setlength{\tabcolsep}{4pt}
\caption{A compact reporting template for automatic T2I verifier selection.}
\label{tab:selection-card}
\begin{tabular}{p{3.2cm}p{8.2cm}}
\toprule
Field & What to report \\
\midrule
Benchmark target & Ability, \longwang{object} definitions, prompt templates, target values. \\
Verifier & Model/checkpoint, version, provider, decoding settings, detector weights. \\
Verifier-use protocol & Prompt, \longwang{object} card, few-shot examples, thresholds, output schema, parser. \\
Scoring rule & Per-image pass rule, treatment of invalid/rejected images, aggregation. \\
Human audit & Annotation categories, count rules, split size, inter-annotator or adjudication procedure. \\
Aggregate reliability & Human score, automatic score, $A-H$, confidence intervals. \\
Generator comparison & Distortion $D$ and its paired interval for each comparison the benchmark should resolve. \\
Image-level reliability & Agreement, Cohen's $\kappa$, valid-image count fidelity. \\
Reject behavior & Reject precision, recall, F1, and auto-pass rate on human-rejected images. \\
Bias diagnostics & \longwang{object}-level MAE, signed bias, $|\mathrm{bias}|/\mathrm{MAE}$. \\
Operational details & Output-compliance rate, compute/API cost, support-set size and tuning time. \\
\bottomrule
\end{tabular}
\end{table}

\input{camera_ready_updates}

\end{document}

%% file: authors.tex
\author{%
  Shurun Li$^{1}$\quad Michael Hua Wang$^{1}$\quad Haibo Zeng$^{2}$\quad Long Wang$^{1}$\\
  \mdseries $^{1}$REASONS Lab, Institute for Network Sciences and Cyberspace, Tsinghua University\\
  \mdseries $^{2}$Virginia Tech\\
  \mdseries\footnotesize\texttt{\{lisr22,wanghua24\}@mails.tsinghua.edu.cn}, \texttt{hbzeng@vt.edu}, \texttt{longwang@tsinghua.edu.cn}%
}

%% file: ack.tex
\begin{ack}
\label{sec:funding}
This work was supported by the National Natural Science Foundation of China (NSFC) under Grant 62132009. Long Wang is the corresponding author. The authors declare no competing interests.
\end{ack}

%% file: tables/app_overall_current.tex
\begin{table}[h]
\centering\scriptsize
\setlength{\tabcolsep}{2.6pt}
\caption{Current original-input open-MLLM results on the balanced 2,280-image split. Human all-image score is 79.30\%. Count out is numeric-output coverage over all images; Count valid uses all human-valid images, counting rejection/null as failure.}
\label{tab:overall-results}
\begin{tabular}{llrrrrrr}
\toprule
Model & Protocol & Count out \% & A all \% & $A-H$ pp & Count valid \% & Agree \% & $\kappa$ \\
\midrule
Gemma3n-E4B & zero count & 100.00 & 83.46 & +4.17 & 88.88 & 82.85 & 0.436 \\
Gemma3n-E4B & zero reject/count & 100.00 & 84.96 & +5.66 & 90.27 & 84.17 & 0.464 \\
Gemma3n-E4B & few-shot count & 100.00 & 82.46 & +3.16 & 87.97 & 81.67 & 0.408 \\
Gemma3n-E4B & few-shot reject/count & 100.00 & 82.50 & +3.20 & 87.73 & 81.36 & 0.398 \\
Gemma4-E4B & zero count & 100.00 & 70.53 & -8.77 & 73.98 & 71.32 & 0.245 \\
Gemma4-E4B & zero reject/count & 100.00 & 69.34 & -9.96 & 73.69 & 71.01 & 0.250 \\
Gemma4-E4B & few-shot count & 100.00 & 76.45 & -2.85 & 78.92 & 74.87 & 0.272 \\
Gemma4-E4B & few-shot reject/count & 99.74 & 76.45 & -2.85 & 78.53 & 75.04 & 0.277 \\
Qwen3.5-27B & zero count & 100.00 & 85.66 & +6.36 & 96.26 & 89.43 & 0.637 \\
Qwen3.5-27B & zero reject/count & 95.13 & 81.89 & +2.59 & 94.68 & 91.27 & 0.721 \\
Qwen3.5-27B & few-shot count & 100.00 & 85.26 & +5.96 & 96.50 & 89.91 & 0.656 \\
Qwen3.5-27B & few-shot reject/count & 91.80 & 78.25 & -1.05 & 93.24 & 92.11 & 0.764 \\
Qwen3.5-9B & zero count & 100.00 & 84.74 & +5.44 & 94.63 & 88.33 & 0.606 \\
Qwen3.5-9B & zero reject/count & 95.61 & 80.88 & +1.58 & 93.44 & 90.53 & 0.703 \\
Qwen3.5-9B & few-shot count & 100.00 & 84.82 & +5.53 & 94.39 & 87.98 & 0.594 \\
Qwen3.5-9B & few-shot reject/count & 96.71 & 81.27 & +1.97 & 94.01 & 90.13 & 0.688 \\
Qwen3.5-4B & zero count & 100.00 & 83.03 & +3.73 & 93.15 & 87.06 & 0.578 \\
Qwen3.5-4B & zero reject/count & 97.50 & 81.62 & +2.32 & 93.82 & 89.34 & 0.661 \\
Qwen3.5-4B & few-shot count & 100.00 & 84.39 & +5.09 & 94.63 & 89.21 & 0.639 \\
Qwen3.5-4B & few-shot reject/count & 95.13 & 80.96 & +1.67 & 94.01 & 91.67 & 0.738 \\
\bottomrule
\end{tabular}
\end{table}

%% file: tables/app_conditional_errors_current.tex
\begin{table}[h]
\centering\scriptsize
\setlength{\tabcolsep}{4pt}
\caption{Conditional numeric errors in the current main matrix. Coverage is the fraction of all 2,087 human-valid images with a numeric prediction. MAE, signed bias and $\rho=|\mathrm{bias}|/\mathrm{MAE}$ condition on those numeric predictions across all seven tasks.}
\label{tab:current-numeric-errors}
\begin{tabular}{llrrrr}
\toprule
Model & Protocol & Numeric coverage \% & MAE & Signed bias & $\rho$ \\
\midrule
Gemma3n-E4B & zero count & 100.00 & 0.1140 & -0.0441 & 0.387 \\
Gemma3n-E4B & zero reject/count & 100.00 & 0.0997 & -0.0278 & 0.279 \\
Gemma3n-E4B & few-shot count & 100.00 & 0.1222 & -0.0446 & 0.365 \\
Gemma3n-E4B & few-shot reject/count & 100.00 & 0.1241 & -0.0561 & 0.452 \\
Gemma4-E4B & zero count & 100.00 & 0.2669 & -0.1260 & 0.472 \\
Gemma4-E4B & zero reject/count & 100.00 & 0.2693 & -0.1476 & 0.548 \\
Gemma4-E4B & few-shot count & 100.00 & 0.2262 & -0.1083 & 0.479 \\
Gemma4-E4B & few-shot reject/count & 99.81 & 0.2295 & -0.1651 & 0.720 \\
Qwen3.5-27B & zero count & 100.00 & 0.0388 & +0.0264 & 0.679 \\
Qwen3.5-27B & zero reject/count & 98.75 & 0.0427 & +0.0194 & 0.455 \\
Qwen3.5-27B & few-shot count & 100.00 & 0.0355 & +0.0259 & 0.730 \\
Qwen3.5-27B & few-shot reject/count & 96.89 & 0.0386 & +0.0277 & 0.718 \\
Qwen3.5-9B & zero count & 100.00 & 0.0565 & +0.0489 & 0.864 \\
Qwen3.5-9B & zero reject/count & 99.19 & 0.0623 & +0.0536 & 0.860 \\
Qwen3.5-9B & few-shot count & 100.00 & 0.0585 & +0.0470 & 0.803 \\
Qwen3.5-9B & few-shot reject/count & 99.52 & 0.0568 & +0.0443 & 0.780 \\
Qwen3.5-4B & zero count & 100.00 & 0.0724 & +0.0244 & 0.338 \\
Qwen3.5-4B & zero reject/count & 99.86 & 0.0648 & +0.0168 & 0.259 \\
Qwen3.5-4B & few-shot count & 100.00 & 0.0561 & +0.0494 & 0.880 \\
Qwen3.5-4B & few-shot reject/count & 98.90 & 0.0518 & +0.0431 & 0.832 \\
\bottomrule
\end{tabular}
\end{table}

%% file: camera_ready_updates.tex
\section{Evidence Updates and Additional Analyses}
\label{app:camera-ready-updates}

This appendix reports the refreshed main counting matrix, the reference and boundary analyses, the prompt-rendering ablation, and the spatial and support-set reruns under the current input policy. The main results, all figures and the overall tables use the refreshed matrix.

\subsection{Matched human-reference audit}

The broad H2 audit and the full traffic-light audit comprise 640 context-specific records for 600 unique images, balanced 300/300 across generators. Forty image identities overlap; three have differing context labels. The canonical matched analysis uses the traffic-light-context H2 labels for those overlaps and retains the original context-specific records for historical reproducibility.

H3 independently annotates the same 600 images with randomized IDs and stripped image metadata. H3 sees the image, task rules and requested target count, but not prior human labels, verifier outputs, generator identity, source filenames or labeled support examples. This is a label- and source-blind audit, not a target-blind audit. All three annotators are paper authors.

\input{tables/app_human_reliability.tex}

The deterministic three-person consensus resolves 594/600 images; six majority-valid images lack two identical counts and remain unresolved. A hybrid reference substitutes resolved consensus for H1 on the packet and retains H1 outside it, excluding the same six unresolved IDs from both comparisons, yielding 2,274 images. On that matched population the human score changes by $+1.01\pp$. The 594-image resolved packet and the 2,274-image hybrid population are distinct estimands.

On the refreshed matrix, mean $\kappa$ over the 20 verifier--protocol configurations changes from $0.539$ to $0.563$ under the hybrid reference: paired difference $+0.024$, with 95\% interval $[0.014,0.034]$. On the resolved packet alone the corresponding difference is $+0.052$ with interval $[0.031,0.076]$. These intervals use 10,000 fixed-generator--cell paired image bootstrap samples, carrying all configurations and both reference judgments together. They quantify reference sensitivity on the two declared populations.

\subsection{Execution and denominator corrections}

The original Gemma counting and spatial configurations performed author-side RGB conversion and Lanczos longest-edge-384 resizing. The replacement protocol passes the original images to each model's native processor, preserving native resize and patch behavior. Query and support images follow the same declared input policy. Historical results and new reruns are versioned separately.

The current native-interleaved Qwen counting runs use original PNG inputs, BF16, no quantization, ordered multimodal content and schema constraints. Qwen3.5-4B/9B use SGLang 0.5.6.post2 and Qwen3.5-27B uses 0.5.16. Model-size contrasts among these runs therefore also reflect runtime differences. Historical-to-refreshed contrasts, including the Gemma contrasts below and the Qwen3.5-27B spatial rerun (two A6000 GPUs with tensor parallelism rather than the historical single device), combine the input-policy change with runtime differences and are not attributed to resizing alone.

The prompt-rendering ablation isolates the multimodal rendering route. On a frozen subset of 228 images, two from each of 114 cells, the same Qwen3.5-9B checkpoint (BF16, original PNGs, temperature 0) receives identical images, support examples, support order, prompts and Human-1 labels under four conditions. vLLM 0.20's default route places the support image blocks ahead of the text, separating each support image from its label; the interleaved string route restores that adjacency, and the explicit OpenAI-content route is token-identical to SGLang and the native Hugging Face processor. Few-shot support lowers $\kappa$ only under the default route (Table~\ref{tab:renderer-ablation}); relative to it, the interleaved string route changes the zero-to-few-shot $\kappa$ effect by $+0.396$ with 95\% interval $[0.230,0.568]$. The result is scoped to this reject/count subset.

\input{tables/app_renderer_ablation.tex}

Count fidelity is measured over all human-valid images, with rejection and nonnumeric output treated as failures. MAE and signed count bias condition on numeric outputs and require numeric coverage alongside them. All-image agreement and numeric-count recovery are reported separately, since higher agreement need not reflect better count recovery.

The refreshed main matrix contains 45,600 judgments: 18,240 new Gemma judgments and 27,360 preserved original-input Qwen judgments. Every Gemma configuration covers all 2,280 images, with no operational or schema failures. In the paired historical-to-original-input Gemma contrast, all eight agreement and $\kappa$ point estimates decrease: agreement changes range from $-1.75$ to $-0.26\pp$, and $\Delta\kappa$ from $-0.052$ to $-0.011$. All eight paired pooled-$A-H$ change intervals on the fixed full suite include zero. The reliability profile is therefore tied to the declared configuration and should be reported with it.

\input{tables/app_current_main_refresh.tex}

\subsection{Generator comparisons and uncertainty}

Section~\ref{sec:verifiers-protocols-metrics} defines the comparison distortion $D$, an effect-size diagnostic that distinguishes an accurate pooled score from preservation of generator differences. Table~\ref{tab:distortion-current} reports $D$ for the Qwen-Image-minus-FLUX difference in all 20 configurations of the main matrix.

\input{tables/app_distortion_current.tex}

For practical use, first declare the generator difference the benchmark is intended to resolve and the comparison error that would be tolerable. Use a representative human-audited sample separate from protocol or threshold selection, estimate the reliability profile and paired intervals on the same images, and compare the plausible distortion and generator-gap uncertainty with that tolerance. Report an unresolved comparison when the evidence cannot support the desired distinction; adequacy is judged against the declared tolerance rather than a universal $\kappa$ threshold.

The refreshed Gemma comparison also separates pooled calibration from preservation of a generator difference. For Gemma 3n few-shot validity-aware counting, the original-input-minus-historical pooled-$A-H$ change is $-0.18\pp$ with paired 95\% interval $[-0.83,0.48]\pp$, while the corresponding change in $D$ is $-2.63\pp$ with interval $[-4.04,-1.32]\pp$. Both use the fixed full suite.

Fixed-cell paired bootstrap resamples image clusters within 114 generator--cell strata (57 cells per generator), carrying all judgments on a sampled image together. Hierarchical sensitivity first resamples structural cells. Excluding the single traffic-light task from both generators leaves 48 cells and 1,920 images. Fixed-cell intervals support inference about the fixed suite, and hierarchical intervals about the broader cell population.

\subsection{Real-image counting and scope}

On 288 official FSC-147 JPEG images~\citep{fsc147}, the audit crosses Gemma 3n E4B-it and Qwen3.5-4B/9B/27B with four protocols, giving 16 conditions. ZC/ZV/FC/FV denote zero-shot count-only, zero-shot validity-aware, few-shot count-only and few-shot validity-aware, respectively; the same abbreviations apply to the three-generator tables below. Conditional numeric errors are accompanied by coverage: $97.53\%$, $96.81\%$ and $92.64\%$ in count bins 7--10, 11--20 and 21--50.

\input{tables/app_fsc_joint.tex}

A same-category, exact-width sensitivity forms 26 low/mid/high triplets (78 images). Its low-minus-high exact-count gap is $32.45\pp$ with interval $[22.12,43.51]\pp$; its high-minus-low relative-MAE increase is $8.79\pp$ with interval $[6.07,11.58]\pp$. This sensitivity controls category and one geometric attribute; density, instance size, occlusion and other natural-image covariates remain uncontrolled, so count is not fully separated from them.

Section~\ref{subsec:rq3} summarizes the strict three-generator analysis, which uses a common Qwen3.5-4B BF16 original-input configuration; Z-Image has H1 labels only. Tables~\ref{tab:strict-threegen-metrics} and~\ref{tab:strict-threegen-pairs} report the per-generator metrics and the pairwise effects. Removing traffic lights changes both point orderings to Qwen-Image above FLUX above Z-Image, so the reversal holds on the fixed suite while the ordering over a broader cell population remains unresolved.

\input{tables/app_strict_threegen_metrics.tex}
\input{tables/app_strict_threegen_pairs.tex}

The spatial extension uses 400 images from 100 frozen GenEval-position prompt clusters, two replicas per generator, with Human-1 labels; 390 have a determinate label. All five open verifiers were rerun with original image inputs under four protocols, giving 8,000 outputs (Table~\ref{tab:spatial-current}). The three Qwen3.5 models stay within $2.6\pp$ of the Human score under every protocol, with agreement of 94.87--96.41\%. Gemma 4 depends on the protocol, from 74.62\% agreement under holistic questioning to 91.28\% under the decomposed protocol, and Gemma 3n over-passes by $12.31$--$21.79\pp$. No configuration reverses the Human ordering of the generators (Qwen-Image minus FLUX: $-23.13\pp$), but Gemma 3n under every protocol and Gemma 4 under holistic questioning shrink that difference by $10.5$--$16.4\pp$. The extension covers one position predicate and is reported separately from the structural-counting estimand.

\input{tables/app_spatial_current.tex}

\subsection{Support-set ablation}

The support-set ablation uses 120 fixed Qwen-Image queries from three tasks and four models with original image inputs. It varies support size (1, 2, 3 and 6 shots, three support-identity replicates each), composition (one valid support replaced by a rejected support) and order. For Qwen3.5-4B/9B, all 45 traffic-light queries among the 120 six-shot queries per replicate exceed the frozen 8,192-token context. Those runs are reported as not executable, so the two Qwen six-shot blocks are excluded whole rather than scored as failures. Count fidelity rises with shot size for Gemma 3n, from 93.16\% at one shot to 97.44\% at six (Table~\ref{tab:support-current}). The practical-stability rule, frozen before inspecting aggregate results, requires the paired six-shot contrast to lie within $\pm3\pp$ and the replicate range to stay within $3\pp$; neither Gemma model meets it through six shots, and it cannot be assessed for the Qwen models. Replacing one valid support with a rejected support, which is possible for the 55 queries whose prompt family has three distinct rejected supports, changes count fidelity by $-2.56$ to $+1.28\pp$, with intervals that include or touch zero; only three queries are human rejects, so reject-rate effects are not estimable. Reversing the support order changes 0.83--4.17\% of predictions overall and 16.67\% of clover predictions for Qwen3.5-9B. The clover label sequence $[3,4,3]$ is palindromic, so that reversal changes only the order of the support images. Intervals use 1,000 task/cell-stratified paired image bootstrap draws.

\input{tables/app_support_current.tex}

\subsection{Detector thresholds and image overlap}

The fixed COCO threshold of $0.25$ is set independently of project labels. The support-trained object detector's 120 support train/validation images have no exact-ID overlap with its 1,080 test images. The support-trained substructure detector uses a separate threshold-calibration split but has 57 train/validation-support IDs among its reported 1,200 test images. Removing them leaves 1,143 images (Table~\ref{tab:detector-disjoint}).

\input{tables/app_detector_disjoint.tex}

%% file: tables/app_human_reliability.tex
\begin{table*}[t]
\caption{Human reliability on the matched 600-image adjudication packet. Binary status collapses \texttt{invalid} and \texttt{clear\_fail} to reject. Exact-count agreement is conditioned on both annotation passes calling an image valid. Intervals are image-bootstrap percentile 95\% CIs (10,000 replicates).}
\label{tab:human-reliability}
\centering
\begin{tabular}{lccc}
\toprule
Pair & Valid/reject agreement (\%) & Cohen's $\kappa$ & Exact count / jointly valid (\%) \\
\midrule
H1-H2 & 90.17 [87.67,92.50] & 0.750 [0.688,0.807] & 408/410 (99.51 [98.75,100.00]) \\
H3-H1 & 92.17 [90.00,94.33] & 0.801 [0.745,0.853] & 415/416 (99.76 [99.25,100.00]) \\
H3-H2 & 95.67 [94.00,97.17] & 0.883 [0.837,0.925] & 433/440 (98.41 [97.16,99.53]) \\
\bottomrule
\end{tabular}
\end{table*}

%% file: tables/app_renderer_ablation.tex
\begin{table}[h]
\centering
\scriptsize
\setlength{\tabcolsep}{4pt}
\caption{Qwen3.5-9B prompt-rendering ablation: reject/count protocol on 228 images (two from each of 114 cells), with the same checkpoint, images, support examples, support order and prompts in every condition. $\Delta\kappa$ is few-shot minus zero-shot; intervals use 10,000 cell-paired bootstrap draws.}
\label{tab:renderer-ablation}
\begin{tabular}{llrrrr}
\toprule
Renderer & Support images & $\kappa$ zero-shot & $\kappa$ few-shot & $\Delta\kappa$ [95\% CI] & Few-shot agree. (\%) \\
\midrule
vLLM 0.20 default (auto$\to$string) & before the text & 0.691 & 0.334 & $-0.357$ $[-0.513,-0.203]$ & 76.75 \\
vLLM 0.20 string, interleaved & next to labels & 0.722 & 0.761 & $+0.039$ $[-0.078,0.147]$ & 93.42 \\
vLLM 0.20 OpenAI content & next to labels & 0.757 & 0.797 & $+0.039$ $[-0.041,0.125]$ & 94.30 \\
SGLang & next to labels & 0.740 & 0.805 & $+0.065$ $[-0.018,0.151]$ & 94.30 \\
\bottomrule
\end{tabular}
\end{table}

%% file: tables/app_current_main_refresh.tex
\begin{table*}[t]
\caption{Refreshed original-input Gemma main counting results (2,280 images per row; Human benchmark score 79.30\%). Query and support images enter the native processor without the author-side longest-edge-384 resize. $\Delta$ denotes the paired original-input-minus-historical-LT384 point difference. $D$ is automatic minus Human distortion of the Qwen-Image--FLUX difference, in pp. ZC/ZV/FC/FV denote zero-/few-shot count-only/reject-count protocols.}
\label{tab:current-main-refresh}
\centering
\small
\resizebox{\textwidth}{!}{%
\begin{tabular}{llrrrrrrr}
\toprule
Model & Prot. & $A-H$ (pp) & Agree. (\%) & $\kappa$ & Count fid. (\%) & $\Delta$Agree. (pp) & $\Delta\kappa$ & $D$ (pp) \\
\midrule
Gemma 3n E4B-it & ZC & +4.17 & 82.85 & 0.436 & 88.88 & -1.23 & -0.045 & -18.33 \\
Gemma 3n E4B-it & ZV & +5.66 & 84.17 & 0.464 & 90.27 & -1.23 & -0.044 & -17.11 \\
Gemma 3n E4B-it & FC & +3.16 & 81.67 & 0.408 & 87.97 & -1.23 & -0.038 & -20.88 \\
Gemma 3n E4B-it & FV & +3.20 & 81.36 & 0.398 & 87.73 & -1.67 & -0.052 & -22.02 \\
Gemma 4 E4B-it & ZC & -8.77 & 71.32 & 0.245 & 73.98 & -1.05 & -0.020 & -20.00 \\
Gemma 4 E4B-it & ZV & -9.96 & 71.01 & 0.250 & 73.69 & -1.75 & -0.041 & -19.04 \\
Gemma 4 E4B-it & FC & -2.85 & 74.87 & 0.272 & 78.92 & -0.61 & -0.029 & -21.49 \\
Gemma 4 E4B-it & FV & -2.85 & 75.04 & 0.277 & 78.53 & -0.26 & -0.011 & -25.00 \\
\bottomrule
\end{tabular}}
\end{table*}

%% file: tables/app_distortion_current.tex
\begin{table}[h]
\centering
\scriptsize
\caption{Comparison distortion of the Qwen-Image-minus-FLUX.2-dev difference in the main counting matrix (1,140 images per generator). The human difference is $25.44\pp$. Intervals use 10,000 fixed-cell paired bootstrap draws over the 57 cells per generator.}
\label{tab:distortion-current}
\begin{tabular}{llrr}
\toprule
Verifier & Protocol & Automatic difference (pp) & $D$ (pp) [95\% CI] \\
\midrule
Qwen3.5-27B & Zero count & +13.42 & $-12.02$ $[-13.77,-10.26]$ \\
Qwen3.5-27B & Zero reject/count & +16.75 & $-8.68$ $[-10.61,-6.67]$ \\
Qwen3.5-27B & Few-shot count & +13.68 & $-11.75$ $[-13.51,-10.00]$ \\
Qwen3.5-27B & Few-shot reject/count & +16.32 & $-9.12$ $[-11.14,-7.19]$ \\
\addlinespace[2pt]
Qwen3.5-9B & Zero count & +11.05 & $-14.39$ $[-16.23,-12.46]$ \\
Qwen3.5-9B & Zero reject/count & +14.56 & $-10.88$ $[-12.98,-8.77]$ \\
Qwen3.5-9B & Few-shot count & +10.53 & $-14.91$ $[-16.84,-12.98]$ \\
Qwen3.5-9B & Few-shot reject/count & +13.60 & $-11.84$ $[-13.86,-9.74]$ \\
\addlinespace[2pt]
Qwen3.5-4B & Zero count & +11.84 & $-13.60$ $[-15.61,-11.58]$ \\
Qwen3.5-4B & Zero reject/count & +14.12 & $-11.32$ $[-13.42,-9.21]$ \\
Qwen3.5-4B & Few-shot count & +13.16 & $-12.28$ $[-14.21,-10.35]$ \\
Qwen3.5-4B & Few-shot reject/count & +17.89 & $-7.54$ $[-9.56,-5.53]$ \\
\addlinespace[2pt]
Gemma3n-E4B & Zero count & +7.11 & $-18.33$ $[-20.88,-15.79]$ \\
Gemma3n-E4B & Zero reject/count & +8.33 & $-17.11$ $[-19.56,-14.74]$ \\
Gemma3n-E4B & Few-shot count & +4.56 & $-20.88$ $[-23.33,-18.51]$ \\
Gemma3n-E4B & Few-shot reject/count & +3.42 & $-22.02$ $[-24.47,-19.56]$ \\
\addlinespace[2pt]
Gemma4-E4B & Zero count & +5.44 & $-20.00$ $[-23.25,-16.75]$ \\
Gemma4-E4B & Zero reject/count & +6.40 & $-19.04$ $[-22.37,-15.70]$ \\
Gemma4-E4B & Few-shot count & +3.95 & $-21.49$ $[-24.47,-18.51]$ \\
Gemma4-E4B & Few-shot reject/count & +0.44 & $-25.00$ $[-27.98,-22.11]$ \\
\bottomrule
\end{tabular}
\end{table}

%% file: tables/app_fsc_joint.tex
\begin{table*}[t]
\caption{Joint FSC-147 scale audit over 16 verifier conditions. Each bin contains 96 official JPEG images; bootstrap resamples images while carrying all conditions for an image together. Exact and within-1 accuracy use all outputs, counting rejects and other nonnumeric outputs as failures. Relative MAE and signed bias use only numeric outputs; numeric coverage reports the fraction retained. Relative MAE divides absolute numeric error by the reference count. All values except signed bias (counts) are percentages.}
\label{tab:fsc-joint}
\centering
\resizebox{\textwidth}{!}{%
\begin{tabular}{lrrrrr}
\toprule
Count bin & Exact [95\% CI] & Within 1 [95\% CI] & Numeric coverage [95\% CI] & Rel. MAE [95\% CI] & Signed bias [95\% CI] \\
\midrule
7-10 & 48.11 [43.55,52.54] & 83.98 [80.53,87.24] & 97.53 [96.35,98.57] & 8.24 [6.75,10.46] & 0.11 [-0.05,0.33] \\
11-20 & 24.87 [20.70,29.10] & 57.49 [53.06,61.78] & 96.81 [95.70,97.85] & 10.27 [9.41,11.18] & -0.46 [-0.72,-0.21] \\
21-50 & 10.68 [7.29,14.19] & 19.79 [16.15,23.76] & 92.64 [90.62,94.60] & 16.58 [15.15,18.13] & -2.90 [-3.63,-2.21] \\
\bottomrule
\end{tabular}}
\end{table*}

%% file: tables/app_strict_threegen_metrics.tex
\begin{table*}[t]
\caption{Strict homogeneous three-generator sensitivity using local SGLang 0.5.6.post2, Qwen3.5-4B, BF16, no quantization, and original 1024-PNG inputs. Each generator has 1,140 images. A dash denotes a count-only protocol that cannot reject.}
\label{tab:strict-threegen-metrics}
\centering
\resizebox{\textwidth}{!}{%
\begin{tabular}{llrrrrrrr}
\toprule
Protocol & Generator & $H$ (\%) & $A$ (\%) & $A-H$ (pp) & Agree. (\%) & $\kappa$ & Count fid. (\%) & Reject P/R/F1 (\%) \\
\midrule
ZC & Qwen & 92.02 & 88.95 & -3.07 & 91.67 & 0.517 & 93.75 & -- \\
ZC & Z-Image & 71.32 & 72.72 & +1.40 & 91.05 & 0.778 & 91.64 & -- \\
ZC & FLUX & 66.58 & 77.11 & +10.53 & 82.46 & 0.572 & 92.47 & -- \\
\addlinespace[2pt]
ZV & Qwen & 92.02 & 88.68 & -3.33 & 91.93 & 0.539 & 93.57 & 57.1/11.1/18.6 \\
ZV & Z-Image & 71.32 & 71.49 & +0.18 & 91.75 & 0.798 & 91.45 & 56.7/22.4/32.1 \\
ZV & FLUX & 66.58 & 74.56 & +7.98 & 86.75 & 0.684 & 94.10 & 100.0/31.8/48.3 \\
\addlinespace[2pt]
FC & Qwen & 92.02 & 90.96 & -1.05 & 94.39 & 0.640 & 96.11 & -- \\
FC & Z-Image & 71.32 & 71.40 & +0.09 & 91.84 & 0.800 & 92.29 & -- \\
FC & FLUX & 66.58 & 77.81 & +11.23 & 84.04 & 0.609 & 92.98 & -- \\
\addlinespace[2pt]
FV & Qwen & 92.02 & 89.91 & -2.11 & 93.68 & 0.616 & 94.84 & 35.0/19.4/25.0 \\
FV & Z-Image & 71.32 & 68.68 & -2.63 & 91.05 & 0.787 & 88.82 & 46.1/53.9/49.7 \\
FV & FLUX & 66.58 & 72.02 & +5.44 & 89.65 & 0.758 & 93.08 & 89.0/51.6/65.3 \\
\bottomrule
\end{tabular}}
\end{table*}

%% file: tables/app_strict_threegen_pairs.tex
\begin{table*}[t]
\caption{Pairwise strict three-generator effects. Full-suite fixed-cell intervals use 20,000 paired bootstrap draws. The final column reports the fixed-cell distortion with traffic lights excluded. The Qwen--FLUX Human point estimate is 25.44 pp, with the canonical 100,000-draw interval $[23.07,27.72]$ pp. For Z--F, the fixed-cell $\Delta A$ intervals are $[-7.11,-1.75]$ (ZC), $[-5.96,-0.26]$ (ZV), $[-9.21,-3.68]$ (FC) and $[-6.23,-0.53]$ (FV) pp against $\Delta H$ interval $[1.93,7.54]$ pp, and the bootstrap probability of a ranking reversal is at least 0.98 under every protocol.}
\label{tab:strict-threegen-pairs}
\centering
\resizebox{\textwidth}{!}{%
\begin{tabular}{llrrclc}
\toprule
Prot. & Pair & $\Delta H$ & $\Delta A$ & Full $D$ [95\% CI] & Point inversion & No-traffic $D$ [95\% CI] \\
\midrule
ZC & Q-F & +25.44 & +11.84 & $-13.60$ $[-15.61,-11.58]$ & no & $-4.17$ $[-6.25,-2.19]$ \\
ZC & Q-Z & +20.70 & +16.23 & $-4.47$ $[-6.58,-2.37]$ & no & $-4.79$ $[-6.98,-2.71]$ \\
ZC & Z-F & +4.74 & -4.39 & $-9.12$ $[-11.32,-6.93]$ & yes & $+0.62$ $[-1.67,+2.92]$ \\
\addlinespace[2pt]
ZV & Q-F & +25.44 & +14.12 & $-11.32$ $[-13.42,-9.21]$ & no & $-5.00$ $[-6.98,-3.02]$ \\
ZV & Q-Z & +20.70 & +17.19 & $-3.51$ $[-5.61,-1.40]$ & no & $-4.17$ $[-6.25,-2.08]$ \\
ZV & Z-F & +4.74 & -3.07 & $-7.81$ $[-10.00,-5.61]$ & yes & $-0.83$ $[-3.02,+1.25]$ \\
\addlinespace[2pt]
FC & Q-F & +25.44 & +13.16 & $-12.28$ $[-14.21,-10.35]$ & no & $-3.54$ $[-5.42,-1.77]$ \\
FC & Q-Z & +20.70 & +19.56 & $-1.14$ $[-3.07,+0.88]$ & no & $-0.83$ $[-2.71,+1.04]$ \\
FC & Z-F & +4.74 & -6.40 & $-11.14$ $[-13.33,-9.04]$ & yes & $-2.71$ $[-4.90,-0.62]$ \\
\addlinespace[2pt]
FV & Q-F & +25.44 & +17.89 & $-7.54$ $[-9.56,-5.53]$ & no & $-2.60$ $[-4.38,-0.73]$ \\
FV & Q-Z & +20.70 & +21.23 & $+0.53$ $[-1.58,+2.63]$ & no & $+1.56$ $[-0.52,+3.65]$ \\
FV & Z-F & +4.74 & -3.33 & $-8.07$ $[-10.35,-5.79]$ & yes & $-4.17$ $[-6.35,-1.98]$ \\
\bottomrule
\end{tabular}}
\end{table*}

%% file: tables/app_spatial_current.tex
\begin{table}[h]
\centering
\small
\setlength{\tabcolsep}{4pt}
\caption{Spatial-relation extension with original image inputs: image-level agreement with Human-1 (\%) and Cohen's $\kappa$ on the 390 images with a determinate Human-1 label (Human pass rate 72.31\%). Protocols: holistic binary (VQAScore-style question), task-guided binary, decomposed (object visibility, then relation) and validity-aware (with a reject option). Gemma 4 E4B-it produced ten schema-noncompliant outputs (one decomposed, nine validity-aware), scored as automatic failures as declared before the rerun. The released tables give 95\% intervals from 5,000 prompt-cluster bootstrap draws.}
\label{tab:spatial-current}
\begin{tabular}{lcccccccc}
\toprule
 & \multicolumn{2}{c}{Holistic binary} & \multicolumn{2}{c}{Task-guided binary} & \multicolumn{2}{c}{Decomposed} & \multicolumn{2}{c}{Validity-aware} \\
\cmidrule(lr){2-3}\cmidrule(lr){4-5}\cmidrule(lr){6-7}\cmidrule(lr){8-9}
Verifier & Agr. & $\kappa$ & Agr. & $\kappa$ & Agr. & $\kappa$ & Agr. & $\kappa$ \\
\midrule
Gemma 3n E4B-it & 73.85 & 0.243 & 82.05 & 0.477 & 75.13 & 0.180 & 76.41 & 0.236 \\
Gemma 4 E4B-it & 74.62 & 0.322 & 88.97 & 0.733 & 91.28 & 0.773 & 89.74 & 0.742 \\
Qwen3.5-4B & 94.87 & 0.870 & 95.90 & 0.896 & 95.13 & 0.877 & 95.90 & 0.898 \\
Qwen3.5-9B & 95.90 & 0.896 & 95.64 & 0.891 & 95.38 & 0.884 & 95.90 & 0.898 \\
Qwen3.5-27B & 96.41 & 0.910 & 95.90 & 0.900 & 95.38 & 0.886 & 95.90 & 0.898 \\
\bottomrule
\end{tabular}
\end{table}

%% file: tables/app_support_current.tex
\begin{table}[h]
\centering
\small
\setlength{\tabcolsep}{6pt}
\caption{Support-set size with original image inputs: count fidelity on human-valid queries (\%), mean over three support-identity replicates with the replicate range in parentheses (120 Qwen-Image queries from three tasks). n.e. = not executable within the frozen 8,192-token context; excluded from scoring (see text).}
\label{tab:support-current}
\begin{tabular}{lcccc}
\toprule
Verifier & 1 shot & 2 shots & 3 shots & 6 shots \\
\midrule
Gemma 3n E4B-it & 93.16 (2.56) & 93.73 (0.85) & 95.16 (1.71) & 97.44 (2.56) \\
Gemma 4 E4B-it & 90.31 (1.71) & 89.46 (2.56) & 91.74 (4.27) & 92.31 (4.27) \\
Qwen3.5-4B & 95.16 (5.13) & 94.87 (4.27) & 94.59 (0.85) & n.e. \\
Qwen3.5-9B & 91.74 (0.85) & 89.74 (1.71) & 92.31 (0.00) & n.e. \\
\bottomrule
\end{tabular}
\end{table}

%% file: tables/app_detector_disjoint.tex
\begin{table}[!ht]
\caption{Post-hoc exact-ID-disjoint sensitivity for the support-trained substructure detector.}
\label{tab:detector-disjoint}
\centering
\begin{tabular}{lrrrr}
\toprule
Slice & $n$ & $A-H$ (pp) & Count fid. (\%) & $\kappa$ \\
\midrule
Reported test & 1200 & -0.75 & 94.92 & 0.861 \\
Exclude 57 exact seen IDs & 1143 & -0.79 & 94.84 & 0.862 \\
\bottomrule
\end{tabular}
\end{table}